\documentclass{article}

\PassOptionsToPackage{numbers,sort&compress}{natbib}

\usepackage[main, final]{neurips_2026}

\usepackage[utf8]{inputenc} 
\usepackage[T1]{fontenc}    
\usepackage{hyperref}       
\usepackage{url}            
\usepackage{booktabs}       
\usepackage{amsfonts}       
\usepackage{nicefrac}       
\usepackage{microtype}      
\usepackage{xcolor}         
\usepackage{amsmath}
\usepackage{tabularx}
\usepackage{threeparttable}
\usepackage[most]{tcolorbox}
 
\usepackage{graphicx}
\usepackage{subcaption}

\usepackage{multirow}
\usepackage{float}
\usepackage{graphicx}
\usepackage{bm}
\usepackage[table]{xcolor}
\definecolor{lightcyanbg}{rgb}{0.88, 1.0, 1.0}
\definecolor{lightgreenbg}{rgb}{0.88, 1.0, 0.88}
\definecolor{lightredbg}{rgb}{1.0, 0.88, 0.88}
\definecolor{lightyellowbg}{rgb}{1.0, 1.0, 0.88}
\definecolor{lightpurplebg}{rgb}{0.9, 0.9, 1.0}
\usepackage{pifont}

\usepackage{xspace}
\newcommand{\ours}{GLEN\xspace}
\newcommand{\oursdata}{SG-Ego\xspace}

\newcommand{\ie}{\textit{i}.\textit{e}.}
\newcommand{\eg}{\textit{e}.\textit{g}.}

\DeclareMathOperator*{\mean}{mean}

\title{Learning to Evolve Scenes: Reasoning about Human Activities with Scene Graphs}

\author{%
  Francesca Pistilli \\
  Politecnico di Torino\\
  Torino, Italy \\
  \texttt{francesca.pistilli@polito.it} \\
\And
  Simone Alberto Peirone \\
  Politecnico di Torino\\
  Torino, Italy \\
  \texttt{simone.peirone@polito.it} \\
  \And
  Giuseppe Averta \\
  Politecnico di Torino\\
  Torino, Italy \\
  \texttt{giuseppe.averta@polito.it} \\
}

\begin{document}

\maketitle

\begin{abstract}
Understanding human behavior while interacting with the surrounding world is crucial for many applications of embodied AI. First-person videos are particularly informative for this problem, as they well capture how activities reshape the scene over time. However, existing approaches often rely on implicit visual or language-aligned representations, disregarding structured reasoning over the scene dynamic.
We argue that explicit, compositional and editable representations of human-environment interactions can play a crucial role for rich grounded activity understanding. To this end, we introduce SG-Ego, a large scale annotation set extending Ego4D with spatio-temporal scene graphs, where relations triplets are consolidated over time into explicit time-evolving descriptions of the scene state. 
To reason over this representation, we propose GLEN, a graph-based model that operates over scene graph sequences to both align them with textual actions and model their temporal evolution. In addition, we formulate the activity-conditioned graph-edit forecasting (A-GEF) problem, a novel task that casts scene dynamics as a sequence of structured transformations conditioned on ongoing actions, enabling explicit reasoning about how scenes change over time.
We validate our approach across multiple downstream tasks, spanning retrieval benchmarks as EgoMCQ and EgoCVR, as well as long-horizon reasoning benchmarks as EXPLORE-Bench and the newly introduced A-GEF. GLEN achieves strong results compared to raw video baselines and it excels in reasoning settings, typically addressed only with MLLMs, while enabling controllable and structured predictions of scene dynamics driven by human activities.
We believe our results establish spatio-temporal scene graphs, together with models that reason over them, as strong compositional and interpretable representations for video understanding and potentially beyond. 
Project page with code at \url{https://francescapistilli.github.io/GLEN}. 
\end{abstract}

\section{Introduction}\label{sec:introduction}
Despite rapid progress in video understanding, modeling how scenes evolve under human actions remains a fundamental challenge.
Reasoning over these transformations is crucial for embodied AI systems that have to plan, forecast and act in shared dynamic environments~\cite{anwar2025remembr, fan2025embodied}. 

First-person videos~\cite{ego4d} are particularly suited to support this kind of reasoning, thanks to their interaction-centered focus on how actions reshape the scene status over time. Despite significant interest of the community on this topic~\cite{jia2022egotaskqa, lin2022egocentric, pramanick2023egovlpv2, mangalam2023egoschema}, current approaches typically handle egocentric video understanding similarly to other computer vision tasks, relying on large-scale trainings to learn compact latent representations from raw pixels or dense visual features. While effective in many practical cases, such approaches entangle objects, interactions, and their temporal dynamics into a single embedding space, making it difficult to reason about how and why a scene changes over time. 
Approaches that include textual narrations like~\cite{lin2022egocentric,pramanick2023egovlpv2} move from purely visual to visual and semantic features, but still rely on implicit representations. The core limitation of these works is not merely a lack of semantics, but rather the absence of an explicit, editable and compositional representation of the human-scene interaction. 
Without such structure, it remains challenging to attribute changes in the scene to specific actions or to reason about how the scene evolves under different activities. 

\begin{figure}[t]
\centering
  \includegraphics[width=0.95\columnwidth,trim={0 0 0cm 0},clip]{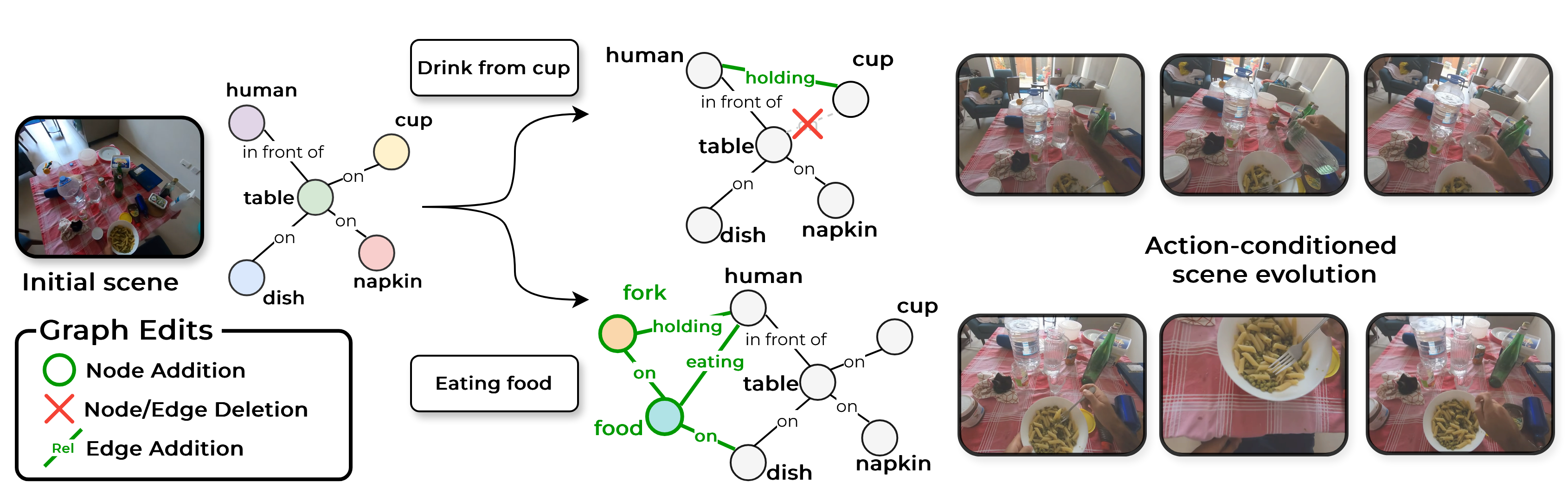}
\caption{\textbf{Reasoning over evolving scenes.} Different activities induce structured changes in a scene over time. We formulate this process as
\emph{activity-conditioned graph edit forecasting} (A-GEF), where the temporal evolution of a scene is modeled as graph edits conditioned on an activity description.}
\label{fig:teaser}
\vspace{-6mm}
\end{figure}

We argue that this can be addressed by structured representations in the form of sequences of scene graphs, where objects and the human are represented as nodes and their functional or spatial relations as edges. This recasts video understanding tasks from a passive feature extraction to an explicit activity-conditioned reasoning over scene structure, with temporal dynamics expressed as structured transformations of scene state rather than implicit shifts in latent embeddings. In other words, sequences of scene graphs provide an interpretable, compositional and editable representation that naturally supports reasoning about \emph{what} changed, \emph{why}, and \emph{how the scene evolves over time}.  \\
While scene graphs have been studied in other contexts like generation from image or video~\cite{ji2020action, cong2021spatial, yang2023panoptic} or future scene anticipation~\cite{nguyen2025hyperglm, peddi2024towards, alliegro2025forescene}, prior works remain fundamentally passive and limited to existing benchmarks, disregarding the activity that generates the scene change. In this work we go beyond existing paradigms and introduce \emph{activity-conditioned graph edit forecasting}, a novel approach for activity understanding that maps the dynamic evolution of a scene conditioned by human activities as edits on a scene graph, see Fig.~\ref{fig:teaser}. 
First, we introduce \oursdata, a large-scale annotation set for Ego4D of spatio-temporal scene graphs, where relational triplets are automatically annotated and then consolidated over time to deliver time-controlled evolving scene descriptions. Then, we propose \ours (Graph-Language Edit Network), a graph neural network trained for graph-language alignment and activity-conditioned graph edits predictions to model scene evolution. 
Finally, we introduce \emph{Activity-conditioned Graph Edit Forecasting} (A-GEF), a novel formulation that models scene dynamics as a sequence of structured transformations over scene graphs conditioned on the ongoing actions, enabling explicit and interpretable reasoning about how scenes evolve over time. 
We validate the quality and completeness of our scene graph annotations on EgoSchema~\cite{mangalam2023egoschema}, showing that scene graph sequences provide rich semantic information. Then, we demonstrate that such representations can be exploited with minimal loss of information compared to raw videos, achieving competitive or state-of-the-art performance on retrieval tasks such as EgoMCQ~\cite{lin2022egocentric} and EgoCVR~\cite{hummel2024egocvr}, a challenging compositional video retrieval benchmark. Finally, \ours achieves state-of-the-art results on reasoning tasks, including EXPLORE-Bench~\cite{yu2026explore} and A-GEF, showing that explicit structured representations can effectively replace or complement black-box MLLMs. \\

Our results highlight how structured representations improve both language grounding and temporal reasoning, while enabling a new class of tasks centered on predicting and manipulating scene dynamics. More broadly, our work advocates for a shift toward structured, explicit representations in egocentric video understanding - moving from describing what is observed to reasoning about how actions transform the scene state.

\section{Related Works}\label{sec:rel}
\paragraph{Video Scene Graph.}
Scene Graph Generation (SGG) aims to represent visual scenes as graphs where objects are the nodes and their relationships are the edges. Recent progress has been driven by transformer-based architectures, which enable end-to-end relational reasoning and have become a dominant paradigm in image-based SGG~\cite{li2022sgtr, im2024egtr}. Alongside architectural advances, substantial effort has addressed the long-tailed distribution of predicates through debiasing and knowledge-aware methods~\cite{chen2023more} and on scaling structured supervision. Synthetic Visual Genome~\cite{park2025synthetic} and its video extension~\cite{gao2026synthetic} enable large-scale extraction of scene graphs from images and videos, improving data availability for relational learning. In parallel, large language models have been leveraged to improve weakly supervised SGG via better triplet extraction and alignment~\cite{kim2024llm4sgg}.
Extending SGG to videos introduces the challenge of modeling temporal dynamics. Action Genome~\cite{ji2020action} formalized video scene graph generation (VidSGG), followed by approaches using spatio-temporal transformers and relational reasoning for dynamic interactions~\cite{cong2021spatial, yang2023panoptic}. Recent work further improves robustness and expressivity through bias mitigation~\cite{nag2023unbiased} and richer structures such as hierarchical and hypergraph representations~\cite{nguyen2024hig, nguyen2025hyperglm}.
Beyond generation, scene graph anticipation methods predict future graphs from past observations, modeling temporal evolution directly in graph space~\cite{nguyen2025hyperglm, peddi2024towards}. However, these approaches remain unconditioned, relying solely on past visual input and failing to model the human activity driving scene changes, limiting controllability over future outcomes.
Recently, scene graphs have been adopted as structured representations for downstream video understanding tasks.
For long-horizon reasoning tasks, 
graph-based representations enable more effective compositional and relational reasoning~\cite{mao2022dynamic, rodin2024action, qiu2025step, zong2025structuring}. Recent approaches further explore graph-based memory and reasoning mechanisms for long-video understanding~\cite{technologies14030150, zemskova2026focusgraph, nguyen2025hyperglm}. 
Despite their success, these methods primarily use scene graphs as auxiliary structures for reasoning or question answering, rather than as a unified representation of scene dynamics. In contrast, we treat scene graphs as a controllable and evolving state, introducing an activity-conditioned formulation where temporal changes are explicitly modeled through graph edit operations.

\paragraph{Graph Edit Models.}
Graph Edit Distance (GED)~\cite{sanfeliu1983distance} is a classical framework for measuring dissimilarity between graphs by computing the minimum-cost sequence of edit operations, as node/edge insertions, deletions, and substitutions, required to transform one graph into another. Since the exact computation of GED is NP-hard, a range of approximations have been proposed, including heuristics strategies~\cite{neuhaus2006fast, riesen2007speeding}, bipartite matching~\cite{riesen2009approximate, serratosa2014fast}, and learning-based methods~\cite{bai2019simgnn, li2019graph, paassen2020graph}.
Recently, Graph Edit Networks (GENs)~\cite{paassen2020graph} have reformulated graph transformation as a learning problem, where neural networks are trained to predict sequences of edit operations between graph pairs. In particular, GENs employ a linear output layer over node representations to produce local edit decisions, including node-level insertions, deletions, and attribute modifications. This formulation enables efficient inference of graph transformations through independent per-node predictions, and leverages learned representations to parameterize edit operations in a data-driven manner.
However, existing GEN formulations primarily address unconditional graph-to-graph transformation, where edits are inferred solely from the source and target graphs.
In contrast, our setting considers conditioned graph evolution, where graph transformations are explicitly influenced by high-level contextual information, requiring the model to adapt its edit decisions based on external conditions.

\section{\oursdata}\label{sec:sgego}
\paragraph{Notation.} We define a spatio-temporal scene graph as a directed multi-edge graph $G_i = (V_i, E_i, \mathbf{X}_i, \mathbf{Y}_i)$, where \(V_i\) is the set of object nodes and \(E_i \subseteq V_i \times V_i\) is the set of directed edges representing pairwise relations. 
Each node \(v \in V_i\) corresponds to an object instance in the scene, to which it is associated a set of attributes \(\mathbf{X}_i \in \mathbb{R}^{|V_i| \times d}\), including its visual features (extracted from the object crop), the corresponding bounding box, and a semantic label describing the category of the object. Each edge \((u, v) \in E_i\) captures a spatial or functional relation from \(u\) to \(v\) and it is associated with a set of attributes \(\mathbf{Y}_i \in \mathbb{R}^{|E_i| \times r}\) comprising the semantic label describing the relation.
Each scene graph \(G_i\) covers a variable temporal window, summarizing all the relations within the window. In particular, we define $G_t$ as the spatial scene graphs associated with frame $t$ and $G_{t:t+T}$ as the spatio-temporal scene graph associated with the temporal window from frame $t$ to $t+T$.\\

\paragraph{Dataset Collection.}
We collect \oursdata, a large scale annotation set of spatio-temporal scene graphs from videos of human activities, based on the Ego4D~\cite{ego4d} dataset, consisting of 3.8M spatio-temporal scene graphs annotated through a \emph{training-free} pipeline. We define a set of $N_{obj}=1480$ node labels and $N_{rel}=387$ relations labels. Each sample spans a temporal window $[t, t+T]$ and is associated with the initial spatial scene graph $G_t$, a spatio-temporal scene graph over the entire temporal window $G_{t:t+T}$ and a textual narration~$act_t$.
For Graph-Text Alignment, we define a training split consisting of 3.8M spatio-temporal scene graphs (\emph{\oursdata Align}).
For A-GEF, we define a training set and a validation set of approximately $360\mathrm{k}$ and $7.2\mathrm{k}$ tuples $(G_{t}, act_{t}, G_{t:t+T})$. We name this split \emph{\oursdata Edit}. These samples are taken from a broad set of videos, following the same scenario distribution of the Ego4D dataset to ensure diversity. 

\subsection{Annotation Pipeline}\label{sec:sgego:anno}
\oursdata implements a \emph{training-free} 
VidSGG strategy divided into three stages. Our pipeline maps a sequence of $T$ frames into a single temporally consistent spatio-temporal scene graphs $G_{t:t+T}$, capturing all the spatial and functional relations in the video. 
We follow a bottom-up approach to extract a rich set of frame-level relations, ground them with a open vocabulary detection model and build a temporally consistent scene graph of the temporal window. Here, we provide a description of the annotation process, see Sec.~\ref{sec:app:sgego} for further implementation details and qualitative visualizations.

\paragraph{Stage 1: Frame-level relations extraction.} Given a video $\mathcal{V}$, we first extract frames at a constant sampling rate~$\tau=5$ fps. Then, we prompt a MLLM (Qwen3.5-9B~\cite{qwen35}) to describe the spatial and functional relations appearing in the frame, avoiding unnecessary details related to the appearance of the object or the environments. We instruct the model to directly produce \emph{(subject\char`_x, relation, object\char`_y)} triplets, where the \emph{\char`_x} and \emph{\char`_y} track the specific instances of the entities. Then, we use rule-based filtering to remove malformed or duplicated triplets. 
For each frame, the captioning pipeline outputs a list of candidate \emph{frame-level} triplets~$\mathcal{T}_i$.  

\paragraph{Stage 2: Frame-level relations grounding.} 
We use GroundingDINO~\cite{liu2024grounding} to ground each candidate frame-level triplet $\mathcal{T}_i$ in the image. Specifically, we concatenate the subject, predicate, and object into a single sentence and feed it to the detector. 
We then extract the queries aligned with the textual spans corresponding to the subject and object, and use them to obtain candidate detections for each entity in the triplet. We use the tracking suffixes from the captions to merge multiple detections from the same object, and apply a set of heuristics to drop duplicates and filter invalid spatial relations (\eg, \emph{to the right}, \emph{inside}, etc...). The resulting objects and relations define the nodes and edges of the frame-level graph, respectively.
From the valid detections, we extract the node attributes consisting of the bounding box and semantic label of the object. 
Objects and relations from the captions are mapped to a fixed set of $N_{obj}=1480$ object and $N_{rel}=387$ classes respectively.
The output of the grounding stage at frame $t$ is a frame-level spatial scene graph $G_t = (V_t, E_t, \mathbf{X}_t, \mathbf{Y}_t)$.

\paragraph{Stage 3: Graph consolidation across frames.}
Each frame-level scene graph provides a partial view of the ongoing interactions in the video, due to occlusions or omitted triplets by the captioner. Given a sequence $G_t,\dots,G_{t+T}$, the consolidation process $\mathcal{C}$ merge them into a single spatio-temporal scene graph that includes all nodes and relations observed during the timespan~\cite{chu2025fine}. Objects correspondences are established via a semantic tracking pipeline based on SAM2~\cite{ravi2025sam} and DINOv2~\cite{oquab2024dinov}.
More specifically, starting from a pair of graphs $G_t$ and $G_{t+1}$, we extract object masks with SAM2 and propagate them from $G_t$ to $G_{t+1}$. We compute $\text{IoU}$ between propagated and detected masks at each frame, and use Hungarian matching to associate tracked and detected objects with $\text{IoU} >0.5$. To improve robustness, we additionally match objects based on DINOv2 feature similarity when tracking fails. The merged graph includes all matched and unmatched nodes and inherits relations from both graphs. This process is repeated sequentially over time to obtain the final consolidated graph~$G_{t:t+T}$.

\section{Method}\label{sec:method}
\begin{figure}[t]
\centering
  \includegraphics[width=1\columnwidth,trim={0.8cm 1cm 0.8cm 1cm},clip]{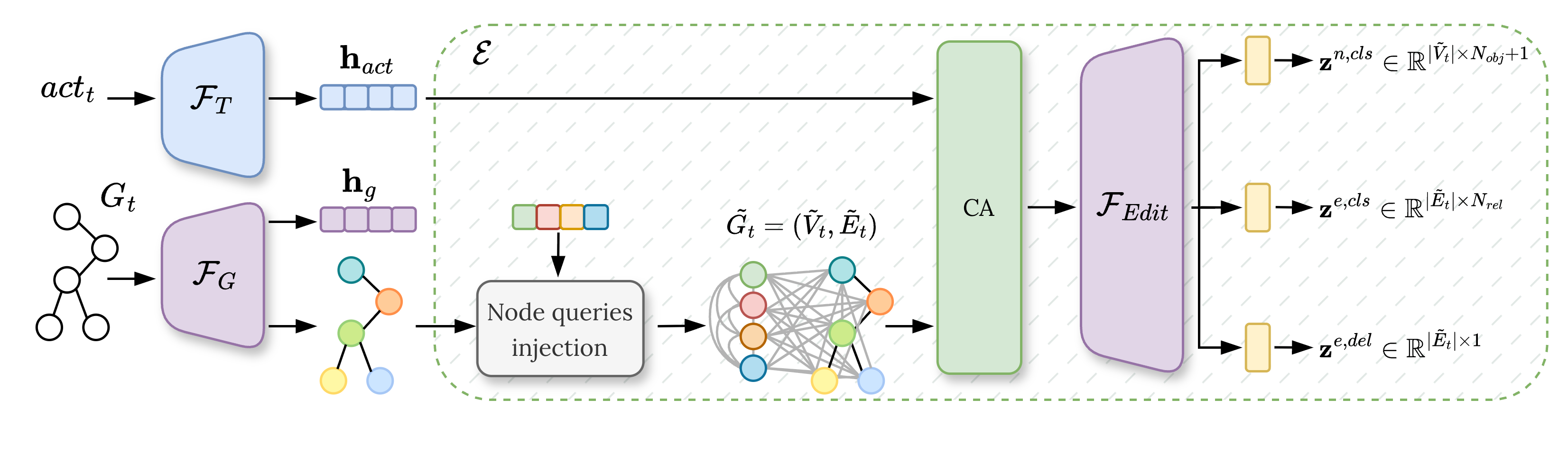}
  \vspace{-0.8cm}
\caption{
\textbf{The \ours architecture for Activity-conditioned Graph Edit Forecasting (A-GEF).}
The model takes a spatial graph $G_t$ and a conditioning action $act_t$ and predicts the necessary edit to evolve the scene graph into the consolidated $G_{t:t+T}$.
For Graph-Text Alignment (GTA), we extract a graph embedding from $G_{t:t+T}$ and align it with the text embedding of $act_t$.
}
\label{fig:model}
\vspace{-3mm}
\end{figure}

We formulate the dynamic evolution of a scene as a novel setting called \emph{activity-conditioned graph edit forecasting} (A-GEF), in which we ask a model to forecast how a spatial scene graph $G_t$ will change based on the activity described in $act_{t}$.
To do this, we consider a \emph{conditional} formulation in which future graph edits are predicted given both the current graph $G_t$ and the upcoming activity $act_t$.
Our approach Graph-Language Edit Network (\ours), see Fig.~\ref{fig:model}, consists of three main components.
The \textbf{Graph Encoder}~$\mathcal{F}_G: \mathcal{G} \mapsto \mathbb{R}^d$ maps from the space of scene graphs $\mathcal{G}_t$ to $d$-dimensional embeddings, while the \textbf{Text Encoder} $\mathcal{F}_T: \mathcal{T} \mapsto \mathbb{R}^d$ maps $act_{t}$ from the text space $\mathcal{T}$ into the same embedding space. 
By leveraging a Graph-Text Alignment (GTA) objective, the model can learn a versatile representation that summarizes the state of the scene and enable spatially grounded reasoning about the objects and relations in the scene.
Finally, the \textbf{Graph Edit Model}~$\mathcal{E}: (\mathcal{G}, \mathcal{T}) \mapsto \mathcal{G}$ defines text-conditioned transformations over the scene graph $\mathcal{G}_t$, allowing structured modifications based on the given activity description. \\
We represent an action as a triplet $(G_t, act_t, G_{t:t+T})$, where $G_t$ is the initial scene graph at timestamp~$t$, $act_t$ is a textual description of the activity starting at $t$, and $G_{t:t+T}$ is the final consolidated scene graph after the activity concludes. Given such triplets, \ours is trained to (i) align the embedding of $G_{t:t+T}$ with the embedding of the action textual description $act_t$ via an alignment objective (GTA) or (ii) transform $G_t$ into the final graph $G_{t:t+T}$ through the Graph Edit Model $\mathcal{E}$ (A-GEF).

\subsection{Graph-Text Alignment}\label{sec:method:align}
We formulate the Graph-Text Alignment task as the problem of learning a joint embedding space in which a spatio-temporal scene graph is aligned with the textual description of the activities occurring within the corresponding temporal window. We train \ours with \emph{Graph-Text Contrastive Alignment (GTCA)} and \emph{Graph-Text Matching (GTM)}.
Our goal is to learn a versatile representation that summarizes the scene graph in a single text-aligned embedding, enabling grounded reasoning about the spatial and functional relations in the scene.

\paragraph{Embeddings extraction.}
We define a Graph Encoder $\mathcal{F}_G: \mathcal{G} \mapsto \mathbb{R}^d$ that maps from the space of scene graphs $\mathcal{G}$ into $d$-dimensional embeddings, and a Text Encoder $\mathcal{F}_T: \mathcal{T} \mapsto \mathbb{R}^d$ that maps the textual narrations into the same embedding space. 
For a scene-graph activity pair $(G_{t:t+T}, act_t)$, we define their corresponding embeddings as $\mathbf{h}_{g}$ and $\mathbf{h}_{act}$. 
$\mathcal{F}_G$ is composed of 
$L$ TripletGCN layers~\cite{Wald2020}, where each layer at depth $l$ updates the node embeddings $\mathbf{x}_j^{(l)}$ through message passing:
\begin{equation}
    \mathbf{x}_j^{(l+1)} = \mathbf{x}_j + \phi_2(\mean_{i \in \text{Neigh}(j)} \phi_1(\mathbf{x}_i || \mathbf{y}_{i\rightarrow j} || \mathbf{x}_j)),
\end{equation}
where $\phi_1$ and $\phi_2$ are MLPs and $\text{Neigh}(j)$ is the set of neighbors of node j.
A similar update rule applies to the edge embeddings $\mathbf{y}_i$. 
By jointly updating node and edge features, the Graph Encoder learns context-aware representations that capture the spatial and semantic structure of the current scene.
We alternate message passing and cross attention layers $CA$ to update the node embeddings $\mathbf{X}^{(l)}$ with the text tokens~$\mathbf{H}_{act}$ from the narrations:
\begin{equation} \label{eq_GE}
    \mathbf{\tilde{X}}^{(l+1)} = LN \left[ \mathbf{X}^{(l+1)} + \alpha \text{CA}(\mathbf{X}^{(l+1)}, \mathbf{H}_{act}) \right],
\end{equation}
where $\text{LN}$ is the layer norm operator and $\alpha$ is a learnable parameter that controls the contribution of the $CA$.
Finally, the encoder outputs a single graph embedding $\mathbf{h}_{g,i}$ by average pooling all the node $\mathbf{X}^{(L)}$ and edge $\mathbf{Y}^{(L)}$ embeddings at the final layer, followed by a MLP projection.
Depending on the task, we enable or disable the cross attention modules.
For Graph-Text Alignment, we train only the Graph Encoder $\mathcal{F}_G$ to align the graph embedding of $G_{t:t+T}$ to the textual embedding of $act_t$.

\paragraph{Graph-Text Contrastive Alignment (GTCA).} 
Given a batch $\mathcal{B} = \{ (G, act)_i \}_{i=1}^{|\mathcal{B}|}$, consisting of~$|\mathcal{B}|$ scene graphs paired with their corresponding textual narrations, we push the model to map close in the embedding space paired graph-text samples, while pushing unpaired samples further apart. We define the batches such that there are at least two samples from the same video to make the model more robust to subtle variations in the scene graphs extracted from the same video.
Formally, we define the graph-to-text contrastive alignment loss~$\mathcal{L}_{g2t}$ as:
\begin{equation}
    \mathcal{L}_{g2t} = -\frac{1}{|\mathcal{B}|}\sum_{i\in\mathcal{B}} 
    \frac{1}{|\mathcal{P}_i|}\sum_{k\in\mathcal{P}_i} \log \frac{ \text{exp} (\mathbf{h}_{g,i}^T \mathbf{h}_{act, k} / \tau)}{\sum_{n\in\mathcal{N}_i} \text{exp} (\mathbf{h}_{g,i}^T \mathbf{h}_{act, n} / \tau)},
\end{equation}

where $\mathcal{P}_i$ is the set of positive samples for the anchor sample $i$, defined as the samples from the batch that share at least one verb and noun labels from the narration $act_i$, and $\mathcal{N}_i$ is the set of negative samples from the same video as sample $i$. We define the symmetric \emph{text-to-graph} contribution $\mathcal{L}_{t2g}$ in a similar way. For GTCA, we disable the Cross Attention layers, keeping the two encoders separate.

\paragraph{Graph-Text Matching (GTM).}
To encourage the network to distinguish subtle differences in the scene graphs, we introduce an additional Graph-Text Matching objective~\cite{li2023blip,pramanick2023egovlpv2}.
Specifically, we mine hard positive pairs and negative pairs from the batch based on the similarity of their graph and text embeddings, enable the CA layers, and feed them to the encoders. 
The resulting embeddings are then passed to an MLP to predict whether the pair is positive or negative.

\subsection{Activity-conditioned Graph Edit Forecasting (A-GEF)}
\paragraph{Problem formulation.}\label{sec:method:problem}
Given a sequence of graphs $\{G_t, G_{t+1}, \dots, G_{t+T}\}$, a Graph Edit Network takes as input a spatial graph $G_t$ and predict a set of edits that transform it into $G_{t+1}$~\cite{paassen2020graph}.
A graph edit function $\delta: \mathcal{G} \rightarrow \mathcal{G}$ is defined as a composition of elementary edit operations which includes node deletion (removing node $i$ and its incident edges), node insertion (adding a new node with attribute $x$), node replacement (updating the attribute of node $i$ to $x$), edge deletion (removing edge $(i,j)$), and edge insertion (adding edge $(i,j)$). These operations form the basis of the graph edit formulation.
In our setting, we extend this formulation to \emph{activity-conditioned graph edit forecasting}. Given the initial scene graph $G_t$ and a textual description of the activity $act_{t}$ starting at timestamp $t$, the goal is to predict the final scene graph $G_{t:t+T}$. Unlike standard graph edit networks~\cite{paassen2020graph}, which model graph evolution purely as a function of the current graph, in our case the dynamic evolution of the graph is conditioned by the activity. 
This conditioning is crucial in our setting, as scene evolution is inherently driven by human actions and the same scene graph will likely evolve differently depending on the activity being performed. 
Differently from the standard formulation, we do not explicitly model node replacement operations. Instead, we restrict the edit space to node deletions and node insertions, and express changes in node attributes through a delete-and-insert~mechanism. 

\paragraph{The Graph Edit Model.}\label{sec:method:edit}
We design a Graph Edit Model~$\mathcal{E}: (\mathcal{G}, \mathcal{T}) \mapsto \mathcal{G}$ that predicts graph transformations explicitly conditioned on the future activity. We first encode the initial graph $G_t$ using the Graph Encoder~$\mathcal{F}_G$ described in Sec.~\ref{sec:method:align}, using Eq. \ref{eq_GE} with $\alpha=0$, obtaining node~$\mathbf{Z}_t^n$ and edge~$\mathbf{Z}_t^e$ embeddings. For the editing objective, the cross-attention layers between node embeddings and text tokens and the graph pooling are disabled; therefore, both node and edge representations correspond directly to the output of the $L$-layer TripletGCN module. 
We then augment the graph structure by adding a fixed number $K$ of learnable node queries, which act as candidate slots for potential future objects. These queries are initialized as learnable embeddings and appended to the set of node features, taking inspiration from~\cite{im2024egtr}. 
Next, we construct a fully connected graph over this augmented node set. Edges involving the newly introduced query nodes and not previously connected old nodes are initialized with noisy features, providing no prior structural bias and allowing the model to freely infer their connectivity and semantics.
After this step, we obtain a noised graph $\tilde{G}_t = (\tilde{V}_t, \tilde{E}_t, \mathbf{\tilde{Z}}^n_t, \mathbf{\tilde{Z}}^e_t)$, with $\tilde{V}_t =|V_t|+K$ nodes and $|\tilde{V}_t|\times|\tilde{V}_t|$~edges.

We condition the augmented graph on the text embedding of the action $\mathbf{h}_{act} = \mathcal{F}_T(act_{t})$ through a Cross-Attention layer, allowing both node and edge features to be conditioned by the action:
\begin{equation}
    \mathbf{\tilde{Z}}_t^{n,cond} = \mathbf{\tilde{Z}}_t^n + CA(\mathbf{\tilde{Z}}_t^n, \mathbf{h}_{act}), \quad\quad \mathbf{\tilde{Z}}_t^{e,cond} = \mathbf{\tilde{Z}}_t^e + CA(\mathbf{\tilde{Z}}_t^e, \mathbf{h}_{act}).
\end{equation}

This allows the model to modulate the entire graph based on the conditioning action, effectively injecting activity-dependent signals into the relational structure.
The conditioned graph is then processed by a second graph encoder, called Graph Edit Encoder $\mathcal{F}_{Edit}$, built as a stack of TripletGCN layers, which propagates the activity-aware information via edge-aware message passing. 
The graph encoder $\mathcal{F}_{Edit}$ maps $\tilde{G}_t$ into a set of graph edits.
Specifically, the node embeddings at the last layer of the model $\mathbf{Z}_t^n$ are fed to a \emph{node update head}, which outputs a object category score $y_i \in \mathbb{R}^{N_{obj}+1}$. This score assigns each node to one of~$N_{obj}$ categories or to a special \emph{no object} class.
Assigning a node from ${G}_t$ to the \emph{no object} class corresponds to a \emph{node deletion}, while assigning a valid class to a query node corresponds to node insertion with a specific node label.
Edge edits are processed through two separate heads to predict the edge deletion score $z_i^{del} \in \mathcal{R}^1$ and its semantic labels $z_i^{cls} \in \mathcal{R}^{N_{rel}}$, possibly more than one. This jointly captures \emph{edge deletion} (by suppressing existing edges) and \emph{edge insertion} (by activating edges involving query nodes with predicted relation labels). 

We train the network by directly supervising graph edit operations. Ground-truth correspondences are established between nodes in $G_{t}$ and $G_{t:t+T}$ using visual features and node labels with a one-to-one mapping, while the additional $K$ query nodes are matched via Hungarian assignment based on label costs. Unmatched predictions are assigned to the \emph{no object} class. Based on these matches, we supervise edge updates and deletions by comparing predicted edges with ground-truth edges between matched node pairs. Given directed multi-edge graphs, edge classification is modeled as a multi-label task over $N_{rel}$ classes per node pair. Edge deletion is formulated as a binary task indicating whether at least one edge exists between two nodes. We use cross-entropy for node classification and binary cross-entropy for edge update and deletion.
\section{Experiments}\label{sec:experiments}
We validate \oursdata and \ours on several benchmarks. On EgoSchema~\cite{mangalam2023egoschema}, we demonstrate that the proposed
scene graph annotations provide a structured and complete representation of the scene that can support long range reasoning about human activities. On EgoMCQ~\cite{lin2022egocentric} and EgoCVR~\cite{hummel2024egocvr}, we show that the graph embeddings produced by \ours are well aligned with the textual description of the corresponding ongoing human activities. Finally, on A-GEF and EXPLORE-Bench, we assess the capability of \ours to predict the dynamic evolution of a scene given an~activity.

\paragraph{Implementation details.}
For the Graph-Text Alignment tasks (EgoMCQ and EgoCVR), we extract the visual and semantic embeddings of nodes and edges using the Perception-Encoder~\cite{bolya2025perception} backbone. For A-GEF and EXPLORE-Bench, we use DINOv2~\cite{oquab2024dinov} for the visual embeddings and MiniLM~\cite{wang2021minilmv2} for the semantic embeddings. We initialize the Text Encoder $\mathcal{F}_T$ from the EgoVLP~\cite{lin2022egocentric} text encoder and keep it frozen. 
We use three and two TripletGCN layers for $\mathcal{F}_G$, and $\mathcal{F}_{Edit}$, respectively. The embedding size is set to $d=512$ for the $\mathcal{F}_G$ and $d=256$ for $\mathcal{F}_{Edit}$. For GTA, we mine $n=3$ negative samples for each anchor in the batch. For A-GEF, we set $K=128$ node queries. Ablations are reported in Sec.~\ref{sec:app:ablations}.
The backbones used to initialize the node and edge embeddings are kept frozen and we train only the Graph Encoders and a small projection head on top of the Text Encoder.
We train our models on a single A100 GPU for approx. 24 hours. 

\subsection{Long-Range Reasoning and Summarization on EgoSchema}
The scene graphs produced by \oursdata capture the relevant spatial and functional interactions in the scene and provide a compact, compositional and editable summary of the ongoing activities happening in the video. 
We validate the quality and completeness of our scene graphs on EgoSchema~\cite{mangalam2023egoschema}, a diagnostic benchmark used to evaluate long-horizon temporal, causal, and intent reasoning over first-person video sequences, by providing a Large Language Model (Qwen3.5-9B~\cite{qwen35}) uniformly sampled scene graphs from the video, results as shown in Table~\ref{tab:egoschema_egomcq}(left). 
Specifically, we convert the graph triplets into textual form and prompt the model to answer questions based solely on the scene state described by these graphs, either using only object and relation labels (\emph{SG-Ego semantic}) or also providing the visual crop of each node (\emph{SG-Ego full}).
Semantic graphs alone already reach 64.6\%, and the full graphs achieve performance close to the model with full access to the raw frames (71.6\% vs. 72.8\%). This is particularly notable since full graphs (\emph{SG-Ego full}) retain only localized object crops rather than the complete visual scene, making questions that require detailed visual recognition challenging. When combined with frames, scene graphs further improve over the video-only setting (74.0\%, +1.2\%). These results show that scene graphs are not a replacement for raw visual input, but rather a structured complementary representation that preserves useful semantic and visual information while explicitly modeling objects and interactions.

\begin{table*}[t]
    \centering
    \caption{\textbf{Long-Range Reasoning on EgoSchema~\cite{mangalam2023egoschema} and Graph-Text Alignment on EgoMCQ~\cite{lin2022egocentric}.} On EgoSchema, we validate the quality and completeness of \oursdata. On EgoMCQ, we show that the graph embeddings from \ours are well-aligned with the caption of the ongoing activities.
    }
    \begin{tabular}{@{\hspace{0.05\textwidth}}p{0.4\textwidth}@{\hspace{0.1\textwidth}}p{0.4\textwidth}@{\hspace{0.05\textwidth}}}
        \begin{minipage}[t]{0.475\textwidth}
            \centering
            \scriptsize
            \setlength{\tabcolsep}{2pt}
            \begin{tabularx}{\linewidth}{Xc}
                \toprule
                {\bf Method} & \textbf{EgoSchema (val) Acc.} \\
                \midrule
                EgoThinker~\cite{pei2025egothinker} & 71.8\\
                \midrule
                Qwen3.5-9B (Blind) & 38.2 \\
                Qwen3.5-9B (Frames) & 72.8 \\
                Qwen3.5-9B (\oursdata semantic) & 64.6 \\
                Qwen3.5-9B (\oursdata full) & 71.6 \\
                Qwen3.5-9B (Frames + \oursdata semantic) & 73.2 \\ 
                \rowcolor{lightgreenbg}
                \textbf{Qwen3.5-9B (Frames + \oursdata full)} & \textbf{74.0}\\ 

                \bottomrule
            \end{tabularx}
        \end{minipage}

         &
        \begin{minipage}[t]{0.4\textwidth}
            \centering
            \scriptsize
            \renewcommand{\arraystretch}{1.06}
            \setlength{\tabcolsep}{2.5pt}
            \begin{tabularx}{\linewidth}{Xccc}
                \toprule
            \multicolumn{1}{c}{\textbf{Method}} & \textbf{Modalities} & \multicolumn{2}{c}{\textbf{EgoMCQ Acc.}} \\
            \cmidrule(lr){3-4} & & \textbf{Inter} & \textbf{Intra} \\
                \midrule
                EgoVLP~\cite{lin2022egocentric} & Video / Text & 90.6 & 57.2 \\
                HierVL~\cite{hiervl} & Video / Text & 90.5 & 52.4 \\
                EgoVLPv2~\cite{pramanick2023egovlpv2} & Video / Text & 91.0 & \textbf{60.9} \\
                HiERO (EgoVLP)~\cite{peirone2025hiero} & Video / Text & \textbf{91.6} & \underline{59.6} \\
                \midrule
                \rowcolor{lightgreenbg}
                \textbf{\ours (Perc. Enc.)}& Graph / Text  & \underline{91.2} & 56.2 \\
                \bottomrule
            \end{tabularx}
        \end{minipage} \\
    \end{tabular}
    \label{tab:egoschema_egomcq}
    \vspace{-5mm}
\end{table*}

\subsection{Graph-Text Alignment on EgoMCQ}
EgoMCQ~\cite{lin2022egocentric} consists of 39k multiple-choice \emph{text-to-video} questions where, given a textual narration, the model has to select the correct video clip between five candidates.
We report Inter-Video accuracy, where clips are from different videos, and Intra-Video accuracy, where they are from the same video in Table~\ref{tab:egoschema_egomcq}(right).
We achieve performance comparable to video-language models \emph{end-to-end} fine-tuned for multimodal contrastive alignment, despite relying on a significantly more compact and structured representation of the video input. 

\subsection{Composed Video Retrieval on EgoCVR}
In Composed Video Retrieval (CVR), the model is tasked to retrieve a video that matches a reference clip after applying a textual modification, \ie, given a source video and a query describing a change in its content, the model must identify the correct video among a set of candidates that matches the modified scene.
We approach this task by directly leveraging the graph–text alignment branch of \ours. Unlike EgoMCQ, which is focused on action-level dynamics, CVR requires more fine-grained reasoning about the compositional structure of the scene and how the instruction query changes it.
Notably, our approach is zero-shot as we do not train or fine-tune for this downstream~task.\\ 
We validate \ours on EgoCVR~\cite{hummel2024egocvr}, a CVR benchmark with 2295 samples, each consisting of a \emph{query clip}, an \emph{instruction} representing the intended modification and a set of \emph{candidate clips} to choose from.
We follow the original EgoCVR formulation, extracting a text embedding from the query clip caption augmented with the modification instructions. Then, we seek for the best match among the available candidates through cosine similarity with their graph embeddings.
To preserve visual consistency between the query and candidate clips, which can otherwise be overshadowed by semantic similarity, we follow TFR-CVR~\cite{hummel2024egocvr} and introduce a preliminary visual filtering step that selects the top-$n_c$ candidates ($n_c=15$) based on similarity to the reference clip. We chose to implement it with CLIP features, ensuring generality without introducing task-specific bias. Results in Table~\ref{tab:egocvr} show that \ours achieves state-of-the-art results, outperforming even methods specifically trained for CVR.
We argue that the scene graph based structure of \ours allows the model to better reason about the spatial and functional relations in the input clips, which is crucial for reasoning about the fine-grained modifications induced by the query.
\begin{table}[!t]
\centering
\setlength\tabcolsep{4.0pt}
\caption{
\textbf{Composed Video Retrieval on EgoCVR}, a retrieval benchmark in which each sample includes a query clip, an instruction with the desired modification, and a set of candidate clips to choose from. We measure performance in terms of Recall@k.
Candidate clips are either from the same (\emph{local}) or different videos (\emph{global}) as the query.
}%
\vspace{2mm}
\scriptsize
{%
\label{tab:egocvr}
\begin{tabular}{lcccccc}
\toprule
    \multicolumn{1}{c}{\textbf{Method}} & \multicolumn{3}{c}{\textbf{Global}} & \multicolumn{3}{c}{\textbf{Local}} \\
    \cmidrule(lr){2-4} \cmidrule(lr){5-7}
     & R@1 & R@5 & R@10 & R@1 & R@2 & R@3 \\
    \midrule
    Random & 0.01 & 0.05 & 0.1 & 25.3 & 38.2 & 50.7 \\
    CLIP & 7.4 & 33.2 & 55.3 & 26.1 & 43.4 & 57.7 \\
    BLIP$_{CoVR}$~\cite{ventura2023covr}$^\dagger$ & 5.4 & 15.2 & 24.3 & 33.1 & 49.5 & 62.9 \\
    BLIP$_{CoVR-ECDE}$~\cite{thawakar2024composed}$^\dagger$ & 6.0 & 14.8 & 24.3 & 33.4 & 49.3 & 63.0 \\
    CIReVL~\cite{cirevl} & 2.0 & 6.8 & 10.2 & 21.6 & 35.1 & 46.0 \\
    TFR-CVR~\cite{hummel2024egocvr} & 14.1 & 39.5 & 54.4 & 44.2 & 61.0 & 73.2 \\
    Thawakar \textit{et al.} ~\cite{thawakar2025beyond}$^\dagger$& {14.6} & \textbf{41.3} & {54.9} & {44.8} & {61.7} & {74.0} \\
    \rowcolor{lightgreenbg}
    \textbf{\ours} & \textbf{16.0} & {40.1} & \textbf{56.9} & \textbf{46.1} & \textbf{64.6} & \textbf{76.7} \\
\bottomrule
\end{tabular}
\begin{tablenotes}
    \centering
    \item \hspace{-1.8cm} $^\dagger$Finetuned on WebVid-CoVR or Dense-WebVid-CoVR for CVR.
\end{tablenotes}
}
\vspace{-3mm}
\end{table}

\subsection{Scene Evolution on A-GEF benchmark}
We evaluate the scene evolution capabilities of \ours on A-GEF built on \oursdata, see Sec.~\ref{sec:method:problem}.
Following standard scene graph evaluation protocols, we evaluate the quality of the predicted graph $\hat{G}_{t:t+T}$ using triplet $Recall@K$ over $(subject, relation, object)$ tuples. We introduce an entropy-based filtering strategy to handle uncertainty in node generation, where we remove nodes for which predicted class entropy exceeds a threshold of 0.5. This reduces noise from ambiguous object assignments and provides a more reliable evaluation on confident predictions.
Following standard scene graph evaluation protocols, we evaluate the quality of the predicted graph $\hat{G}_{t:t+T}$ using triplet $Recall@K$. We introduce an entropy-based filtering strategy to handle uncertainty in node generation, where we remove nodes for which predicted class entropy exceeds a threshold of 0.5. This reduces noise from ambiguous object assignments and provides a more reliable evaluation on confident predictions.
We compare \ours with two families of baselines: a \emph{static} approach that replicates $G_t$ as the $\hat{G}_{t:t+T}$ prediction, and baselines based on Qwen3.5-9B~\cite{qwen35}. For the latter, we provide the model with the list of triplets of the start graph in textual form, the narration of the activity and the lists of possible objects and relations, and ask it to predict the triplets of the target scene. 
\newline We evaluate three settings: zero-shot with text only, zero-shot with the start frame as an additional visual input, and with the start frame after LoRA fine-tuning on the A-GEF training set.
As shown in Table~\ref{tab:agef_validation}, the start frame helps the LLM (15.30 vs. 9.19 R@20) and fine-tuning further improves it (22.70), but all the LLM settings remain below the static baseline. In all settings the LLM predicts few triplets, which leaves the recall almost unchanged at different $K$ values.
\ours consistently outperforms all the baselines, including the fine-tuned LLM (36.69 vs. 22.70 R@20), demonstrating the effectiveness of explicitly modeling activity-conditioned graph edits for capturing scene evolution. Notably, the gap with the static baseline highlights that a large portion of scene dynamics cannot be explained by persistence alone, but requires modeling how objects and relations change under actions. At the same time, the improvement over the LLM baselines, even when trained on the same data, suggests that operating in a structured graph space provides a stronger inductive bias than purely language-driven predictions. Overall, these results validate our formulation of scene evolution as a sequence of activity-conditioned graph transformations, enabling more accurate and interpretable predictions of future scene structure.
\begin{table}[!t]
    \caption{
    \textbf{Action-conditioned Graph Edit Forecasting (A-GEF) on \oursdata}. 
    We compare \ours with Qwen3.5 by providing the model with the textual triplets list and asking it to predict the future triplets, and with a static baseline that replicates $G_{t}$ as the final $\hat{G}_{t:t+T}$ prediction.
    Our approach outperforms both Qwen and the static baseline, demonstrating that graph edits are an effective way to capture scene evolution.
    LoRA: fine-tuned on the A-GEF training set.
    }
    \vspace{2mm}
    \label{tab:agef_validation}
    \centering
    \scriptsize
    \begin{tabular}{lccc}
        \toprule
        \multirow{2}{*}{\bf Method}  & \multicolumn{3}{c}{\bf Triplet Recall}\\ 
        & R@20 & R@50 & R@100 \\
        \midrule
        Qwen3.5-9B (text only)~\cite{qwen35}      & 9.19 & 9.19 & 9.19             \\
        Qwen3.5-9B (text + start frame)~\cite{qwen35}      & 15.30 & 15.30 & 15.30             \\
        Qwen3.5-9B (LoRA, text + start frame)~\cite{qwen35}      & 22.70 & 22.78 & 22.78             \\
        \midrule
        $G_t$ (\emph{static}) & 25.94 & 25.94 & 25.94 \\
        \rowcolor{lightgreenbg}
        \textbf{\ours} & \textbf{36.69} & \textbf{45.39} & \textbf{49.91}\\
        \bottomrule
    \end{tabular}
    \vspace{-3mm}
\end{table}

\subsection{Long Horizon Reasoning on EXPLORE-Bench}
We evaluate the generalization capabilities of our activity-conditioned graph edit forecasting framework on EXPLORE-Bench~\cite{yu2026explore}, a benchmark designed to assess long-horizon reasoning over sequences of actions. The task requires predicting the final scene resulting from the sequential application of atomic actions to an initial state.
We cast this problem within our framework by first extracting a scene graph from the initial scene (Sec.~\ref{sec:sgego:anno}), and conditioning \ours on the sequence of actions to predict the resulting scene graph, which serves as a structured representation of the final state. Notably, we do not train our model on this benchmark. Instead, we directly apply the model trained on A-GEF, demonstrating its ability to transfer to long-horizon reasoning without any task-specific supervision. This is particularly significant because such long-horizon compositional reasoning tasks are typically addressed using MLLMs only. In contrast, we show that a structured, graph-based formulation alone is sufficient to model long multi-step scene evolution under actions. \\
We follow the evaluation protocol of~\cite{yu2026explore}, measuring both object and relation level accuracy. For object evaluation, Sentence-BERT embeddings are used to compute a similarity matrix between predicted and ground-truth objects, yielding a soft coverage score $\bm{S_{obj}}$. For relations, a standardized prompting strategy is used to guide a large language model in scoring predicted relations on a 0–5 scale against ground-truth annotations, resulting in the $\bm{S_{rel}}$ metric.
\newline As shown in Table~\ref{tab:explore}, our method achieves state-of-the-art performance on $\bm{S_{obj}}$ and competitive results on $\bm{S_{rel}}$ compared to MLLMs baselines. 
Since Qwen3.5-9B \cite{qwen35} is used in the first stage of our pipeline (Sec. \ref{sec:sgego:anno}) for scene captioning and description, we also evaluate it directly on EXPLORE-Bench. \ours improves $S_{obj}$ over Qwen3.5-9B by 8.6 points on the full benchmark (65.59 vs. 56.95), with comparable $S_{rel}$ (2.69 vs. 2.68), showing that the gains do not come from the choice of the underlying MLLM, but from explicit reasoning over a structured representation.
These results highlight the strong generalization ability of our structured, activity-conditioned formulation for modeling long-horizon scene evolution, and demonstrate that graph-based reasoning provides a viable alternative to language-based approaches for compositional temporal understanding.
\begin{table}[!t] 
\caption{
\textbf{Long horizon reasoning on EXPLORE-Bench.} 
Starting from an initial frame, the model must predict the final scene after a sequence of actions.
Object predictions are evaluated in terms of recall ($\bm{S_{obj}}$), while relations are evaluated with an LLM-based scoring function ($\bm{S_{rel}}$). $^\dagger$ Reproduced results using the official protocol.
}
\vspace{2mm}
\label{tab:explore}
\centering
\scriptsize
\renewcommand{\arraystretch}{1.1} 
{
\begin{tabular}{l|cc|cc|cc|cc}
\toprule
\multirow{2}{*}{\centering\textbf{Methods}} 
& \multicolumn{2}{c|}{\texttt{Short}} 
& \multicolumn{2}{c|}{\texttt{Medium}} 
& \multicolumn{2}{c|}{\texttt{Long}} 
& \multicolumn{2}{c}{\texttt{Full}} \\

& {$\bm{S_{obj}}$} & {$\bm{S_{rel}}$} 
& {$\bm{S_{obj}}$} & {$\bm{S_{rel}}$} 
& {$\bm{S_{obj}}$} & {$\bm{S_{rel}}$} 
& {$\bm{S_{obj}}$} & {$\bm{S_{rel}}$} \\ 
\midrule

GPT-5.2-Chat & 59.91 & 2.70  & 59.88 & 2.65 & 58.06 & 2.61 & 59.69 & 2.67 \\
Gemini-3-Pro & 61.29 & 2.77 & 60.99 & 2.74 & 59.17 & 2.70 & 60.94 & 2.75\\ 
LLaVA-OneVision-1.5-8B & 53.25 & 2.51 & 51.21 & 2.44 &  47.62 & 2.41 & 51.87 & 2.47 \\
Qwen3-VL-8B-Instruct & 61.34 & 2.84 & 60.78 & \textbf{2.81} & 56.83 & \textbf{2.71} & 60.63 & \textbf{2.82} \\ 
Qwen3-VL-8B-Thinking & 63.77 & \textbf{2.85} & 62.61 & 2.78 & 58.02 & 2.63 & 62.70 & 2.80 \\ 
Qwen3.5-9B-Instruct$^\dagger$& 58.73 & 2.74 & 56.20 & 2.63 & 51.21 & 2.55 &  56.95 &  2.68 \\ 
\rowcolor{lightgreenbg}
\textbf{\ours} & \textbf{66.12} & 2.67 & \textbf{66.71} & 2.73 & \textbf{59.37} & 2.67 & \textbf{65.59} & 2.69 \\

\bottomrule
\end{tabular}%
}
\vspace{-3mm}
\end{table}

\vspace{-2mm}
\section{Future directions}
\vspace{-1mm}
Our formulation models evolving scenes as editable graphs, enabling explicit reasoning over how actions transform the state of the environment. While this is valuable for human activity understanding, a particularly promising direction lies in embodied-AI and especially robotics.
By extending \ours with spatial prediction, it naturally becomes a structured scene transition model particularly suited to let a robotic agent reason not only about \emph{what} changes, but also about \emph{where} such changes occur and \emph{why}. This provides a direct bridge between perception and action, supporting grounded planning and control through explicit predictions of future scene states, and may also serve as rich intermediate representation for visual policy learning trained from graph-based expert demonstrations. 

\vspace{-2mm}
\section{Conclusion}
\vspace{-1mm}
We introduce a new paradigm for egocentric video understanding by modeling scenes as structured graphs and their evolution as activity-conditioned graph edits. This enables explicit reasoning over how actions transform the environment, unifying representation learning, language alignment, and temporal prediction within a single framework. Our results show that operating in structured space improves language grounding and temporal reasoning, while providing interpretability and controllability. More broadly, this work highlights editable, structured representations as a promising foundation bridging perception and action. 

\section*{Acknowledgments}
This study was carried out within the FAIR - Future Artificial Intelligence Research and received funding from the European Union Next-GenerationEU (PIANO NAZIONALE DI RIPRESA E RESILIENZA (PNRR) – MISSIONE 4 COMPONENTE 2, INVESTIMENTO 1.3 – D.D. 1555 11/10/2022, PE00000013). This manuscript reflects only the authors’ views and opinions, neither the European Union nor the European Commission can be considered responsible for them. We acknowledge the CINECA award under the ISCRA initiative, for the availability of high performance computing resources and support. 

{
    \small
    \bibliographystyle{unsrtnat}
    \bibliography{main}

@String(CVPR= {IEEE Conf. Comput. Vis. Pattern Recog.})

@String(ICCV= {Int. Conf. Comput. Vis.})

@String(ECCV= {Eur. Conf. Comput. Vis.})

@String(WACV= {Winter Conf. Appl. of Comput. Vis.})

@String(NIPS= {Adv. Neural Inform. Process. Syst.})

@String(ICLR = {Int. Conf. Learn. Represent.})

@string(ICML = {Int. Conf. Mach. Learning})

@String(AAAI = {AAAI})

@String(CVPR  = {CVPR})

@String(ICCV  = {ICCV})

@String(ECCV  = {ECCV})

@String(WACV  = {WACV})

@String(NIPS  = {NeurIPS})

@String(ICLR  = {ICLR})

@String(ICML  = {ICML})

@inproceedings{ego4d,
  title     = {Ego4d: Around the world in 3,000 hours of egocentric video},
  author    = {Grauman, Kristen and Westbury, Andrew and Byrne, Eugene and Chavis, Zachary and Furnari, Antonino and Girdhar, Rohit and Hamburger, Jackson and Jiang, Hao and Liu, Miao and Liu, Xingyu and others},
  booktitle = CVPR,
  year      = {2022}
}

@article{serratosa2014fast,
  title={Fast computation of bipartite graph matching},
  author={Serratosa, Francesc},
  journal={Pattern Recognition Letters},
  year={2014}
}

@article{riesen2009approximate,
  title={Approximate graph edit distance computation by means of bipartite graph matching},
  author={Riesen, Kaspar and Bunke, Horst},
  journal={Image and Vision computing},
  year={2009}
}

@inproceedings{riesen2007speeding,
  title={Speeding up graph edit distance computation with a bipartite heuristic.},
  author={Riesen, Kaspar and Fankhauser, Stefan and Bunke, Horst},
  booktitle={MLG},
  year={2007}
}

@inproceedings{neuhaus2006fast,
  title={Fast suboptimal algorithms for the computation of graph edit distance},
  author={Neuhaus, Michel and Riesen, Kaspar and Bunke, Horst},
  booktitle={Joint IAPR International Workshops on Statistical Techniques in Pattern Recognition (SPR) and Structural and Syntactic Pattern Recognition (SSPR)},
  year={2006},
}

@inproceedings{li2019graph,
  title={Graph matching networks for learning the similarity of graph structured objects},
  author={Li, Yujia and Gu, Chenjie and Dullien, Thomas and Vinyals, Oriol and Kohli, Pushmeet},
  booktitle={ICML},
  year={2019},
}

@inproceedings{bai2019simgnn,
  title={Simgnn: A neural network approach to fast graph similarity computation},
  author={Bai, Yunsheng and Ding, Hao and Bian, Song and Chen, Ting and Sun, Yizhou and Wang, Wei},
  booktitle={Proceedings of the twelfth ACM international conference on web search and data mining},
  year={2019}
}

@article{alliegro2025forescene,
  title={FORESCENE: FOREcasting human activity via latent SCENE graphs diffusion},
  author={Alliegro, Antonio and Pistilli, Francesca and Tommasi, Tatiana and Averta, Giuseppe},
  journal={arXiv preprint arXiv:2503.06182},
  year={2025}
}

@inproceedings{thawakar2025beyond,
  title={Beyond simple edits: Composed video retrieval with dense modifications},
  author={Thawakar, Omkar and Demidov, Dmitry and Thawkar, Ritesh and Anwer, Rao Muhammad and Shah, Mubarak and Khan, Fahad Shahbaz and Khan, Salman},
  booktitle={ICCV},
  year={2025}
}

@inproceedings{peddi2024towards,
  title={Towards scene graph anticipation},
  author={Peddi, Rohith and Singh, Saksham and Saurabh and Singla, Parag and Gogate, Vibhav},
  booktitle={ECCV},
  year={2024},
}

@article{gao2026synthetic,
  title={Synthetic Visual Genome 2: Extracting Large-scale Spatio-Temporal Scene Graphs from Videos},
  author={Gao, Ziqi and Zhang, Jieyu and Ikezogwo, Wisdom Oluchi and Park, Jae Sung and You, Tario G and Ogbu, Daniel and Zheng, Chenhao and Huang, Weikai and Yang, Yinuo and Han, Winson and others},
  journal={arXiv preprint arXiv:2602.23543},
  year={2026}
}

@inproceedings{park2025synthetic,
  title={Synthetic visual genome},
  author={Park, Jae Sung and Ma, Zixian and Li, Linjie and Zheng, Chenhao and Hsieh, Cheng-Yu and Lu, Ximing and Chandu, Khyathi and Kong, Quan and Kobori, Norimasa and Farhadi, Ali and others},
  booktitle={CVPR},
  year={2025}
}

@inproceedings{kim2024llm4sgg,
  title={Llm4sgg: Large language models for weakly supervised scene graph generation},
  author={Kim, Kibum and Yoon, Kanghoon and Jeon, Jaehyeong and In, Yeonjun and Moon, Jinyoung and Kim, Donghyun and Park, Chanyoung},
  booktitle={CVPR},
  year={2024}
}

@inproceedings{chen2023more,
  title={More knowledge, less bias: Unbiasing scene graph generation with explicit ontological adjustment},
  author={Chen, Zhanwen and Rezayi, Saed and Li, Sheng},
  booktitle={WACV},
  year={2023}
}

@inproceedings{li2022sgtr,
  title={Sgtr: End-to-end scene graph generation with transformer},
  author={Li, Rongjie and Zhang, Songyang and He, Xuming},
  booktitle={CVPR},
  year={2022}
}

@inproceedings{im2024egtr,
  title={Egtr: Extracting graph from transformer for scene graph generation},
  author={Im, Jinbae and Nam, JeongYeon and Park, Nokyung and Lee, Hyungmin and Park, Seunghyun},
  booktitle={CVPR},
  year={2024}
}

@inproceedings{cong2021spatial,
  title={Spatial-temporal transformer for dynamic scene graph generation},
  author={Cong, Yuren and Liao, Wentong and Ackermann, Hanno and Rosenhahn, Bodo and Yang, Michael Ying},
  booktitle={ICCV},
  year={2021}
}

@inproceedings{nguyen2024hig,
  title={Hig: Hierarchical interlacement graph approach to scene graph generation in video understanding},
  author={Nguyen, Trong-Thuan and Nguyen, Pha and Luu, Khoa},
  booktitle={CVPR},
  year={2024}
}

@inproceedings{nguyen2025hyperglm,
  title={Hyperglm: Hypergraph for video scene graph generation and anticipation},
  author={Nguyen, Trong-Thuan and Nguyen, Pha and Cothren, Jackson and Yilmaz, Alper and Luu, Khoa},
  booktitle={CVPR},
  year={2025}
}

@inproceedings{yang2023panoptic,
  title={Panoptic video scene graph generation},
  author={Yang, Jingkang and Peng, Wenxuan and Li, Xiangtai and Guo, Zujin and Chen, Liangyu and Li, Bo and Ma, Zheng and Zhou, Kaiyang and Zhang, Wayne and Loy, Chen Change and others},
  booktitle={CVPR},
  year={2023}
}

@inproceedings{nag2023unbiased,
  title={Unbiased scene graph generation in videos},
  author={Nag, Sayak and Min, Kyle and Tripathi, Subarna and Roy-Chowdhury, Amit K},
  booktitle={CVPR},
  year={2023}
}

@inproceedings{ji2020action,
  title={Action genome: Actions as compositions of spatio-temporal scene graphs},
  author={Ji, Jingwei and Krishna, Ranjay and Fei-Fei, Li and Niebles, Juan Carlos},
  booktitle={CVPR},
  year={2020}
}

@inproceedings{cirevl,
  title={Vision-by-Language for Training-Free Compositional Image Retrieval},
  author={Shyamgopal Karthik and Karsten Roth and Massimiliano Mancini and Zeynep Akata},
  booktitle={ICLR},
  year={2024}
}

@inproceedings{thawakar2024composed,
  title={Composed Video Retrieval via Enriched Context and Discriminative Embeddings},
  author={Thawakar, Omkar and Naseer, Muzammal and Anwer, Rao Muhammad and Khan, Salman and Felsberg, Michael and Shah, Mubarak and Khan, Fahad Shahbaz},
  booktitle={CVPR},
  year={2024}
}

@inproceedings{ventura2023covr,
  title={{CoVR}: Learning Composed Video Retrieval from Web Video Captions},
  author={Ventura, Lucas and Yang, Antoine and Schmid, Cordelia and Varol, G{\"u}l},
  booktitle={AAAI},
  year={2024}
}

@inproceedings{Wald2020,
    title = {{Learning 3D Semantic Scene Graphs from 3D Indoor Reconstructions}},
    author = {Wald, Johanna and Dhamo, Helisa and Navab, Nassir and Tombari, Federico},
    booktitle = {CVPR}, 
    year = {2020}
}

@article{sanfeliu1983distance,
  title={A distance measure between attributed relational graphs for pattern recognition},
  author={Sanfeliu, Alberto and Fu, King-Sun},
  journal={IEEE Transactions on Systems, Man, and Cybernetics},
  year={1983},
}

@inproceedings{paassen2020graph,
  title={Graph edit networks},
  author={Paassen, Benjamin and Grattarola, Daniele and Zambon, Daniele and Alippi, Cesare and Hammer, Barbara},
  booktitle={ICLR},
  year={2021}
}

@article{zemskova2026focusgraph,
  title={FocusGraph: Graph-Structured Frame Selection for Embodied Long Video Question Answering},
  author={Zemskova, Tatiana and Andryushenko, Solomon and Obrubov, Ilya and Khoruzhaia, Viktoriia and Eroshenko, Ekaterina and Derevyanka, Ekaterina and Yudin, Dmitry},
  journal={arXiv preprint arXiv:2603.04349},
  year={2026}
}

@Article{technologies14030150,
AUTHOR = {Linok, Sergey and Semenov, Vadim and Trunova, Anastasia and Bulichev, Oleg and Yudin, Dmitry},
TITLE = {DyGEnc: Encoding a Sequence of Textual Scene Graphs to Reason and Answer Questions in Dynamic Scenes},
JOURNAL = {Technologies},
YEAR = {2026},
}

@inproceedings{anwar2025remembr,
  title={Remembr: Building and reasoning over long-horizon spatio-temporal memory for robot navigation},
  author={Anwar, Abrar and Welsh, John and Biswas, Joydeep and Pouya, Soha and Chang, Yan},
  booktitle={ICRA},
  year={2025},
}

@inproceedings{zong2025structuring,
  title={Structuring Video Semantics with Temporal Triplets for Zero-Shot Video Question Answering},
  author={Zong, Linlin and Zhai, Xinyu and Liu, Xinyue and Liang, Wenxin and Zhang, Xianchao and Xu, Bo},
  booktitle={CIKM},
  year={2025}
}

@inproceedings{rodin2024action,
  title={Action scene graphs for long-form understanding of egocentric videos},
  author={Rodin, Ivan and Furnari, Antonino and Min, Kyle and Tripathi, Subarna and Farinella, Giovanni Maria},
  booktitle={CVPR},
  year={2024}
}

@inproceedings{qiu2025step,
  title={STEP: Enhancing Video-LLMs' Compositional Reasoning by Spatio-Temporal Graph-guided Self-Training},
  author={Qiu, Haiyi and Gao, Minghe and Qian, Long and Pan, Kaihang and Yu, Qifan and Li, Juncheng and Wang, Wenjie and Tang, Siliang and Zhuang, Yueting and Chua, Tat-Seng},
  booktitle={CVPR},
  year={2025}
}

@inproceedings{mao2022dynamic,
  title={Dynamic multistep reasoning based on video scene graph for video question answering},
  author={Mao, Jianguo and Jiang, Wenbin and Wang, Xiangdong and Feng, Zhifan and Lyu, Yajuan and Liu, Hong and Zhu, Yong},
  booktitle={HLT-NAACL},
  year={2022}
}

@inproceedings{fan2025embodied,
  title={Embodied videoagent: Persistent memory from egocentric videos and embodied sensors enables dynamic scene understanding},
  author={Fan, Yue and Ma, Xiaojian and Su, Rongpeng and Guo, Jun and Wu, Rujie and Chen, Xi and Li, Qing},
  booktitle={ICCV},
  year={2025}
}

@inproceedings{hiervl,
  title     = {HierVL: Learning Hierarchical Video-Language Embeddings},
  author    = {Ashutosh, Kumar and Girdhar, Rohit and Torresani, Lorenzo and Grauman, Kristen},
  booktitle = {CVPR},
  year      = {2023}
}

@inproceedings{lin2022egocentric,
  title     = {Egocentric video-language pretraining},
  author    = {Lin, Kevin Qinghong and Wang, Jinpeng and Soldan, Mattia and Wray, Michael and Yan, Rui and XU, Eric Z and Gao, Difei and Tu, Rong-Cheng and Zhao, Wenzhe and Kong, Weijie and others},
  booktitle = NIPS,
  year      = {2022}
}

@inproceedings{pramanick2023egovlpv2,
  title     = {Egovlpv2: Egocentric video-language pre-training with fusion in the backbone},
  author    = {Pramanick, Shraman and Song, Yale and Nag, Sayan and Lin, Kevin Qinghong and Shah, Hardik and Shou, Mike Zheng and Chellappa, Rama and Zhang, Pengchuan},
  booktitle = {ICCV},
  year      = {2023}
}

@inproceedings{mangalam2023egoschema,
  title   = {Egoschema: A diagnostic benchmark for very long-form video language understanding},
  author  = {Mangalam, Karttikeya and Akshulakov, Raiymbek and Malik, Jitendra},
  booktitle = NIPS,
  year    = {2023}
}

@inproceedings{jia2022egotaskqa,
  title   = {Egotaskqa: Understanding human tasks in egocentric videos},
  author  = {Jia, Baoxiong and Lei, Ting and Zhu, Song-Chun and Huang, Siyuan},
  booktitle = NIPS,
  year    = {2022}
}

@article{yu2026explore,
  title={EXPLORE-Bench: Egocentric Scene Prediction with Long-Horizon Reasoning},
  author={Yu, Chengjun and Zhu, Xuhan and Du, Chaoqun and Yu, Pengfei and Zhai, Wei and Cao, Yang and Zha, Zheng-Jun},
  journal={arXiv preprint arXiv:2603.09731},
  year={2026}
}

@inproceedings{hummel2024egocvr,
  title={Egocvr: An egocentric benchmark for fine-grained composed video retrieval},
  author={Hummel, Thomas and Karthik, Shyamgopal and Georgescu, Mariana-Iuliana and Akata, Zeynep},
  booktitle={ECCV},
  year={2024},
}

@misc{qwen35,
  title = {Qwen3.5: Towards Native Multimodal Agents},
  url = {https://qwen.ai/blog?id=qwen3.5},
  year = {2026}
}

@inproceedings{liu2024grounding,
  title={Grounding dino: Marrying dino with grounded pre-training for open-set object detection},
  author={Liu, Shilong and Zeng, Zhaoyang and Ren, Tianhe and Li, Feng and Zhang, Hao and Yang, Jie and Jiang, Qing and Li, Chunyuan and Yang, Jianwei and Su, Hang and others},
  booktitle={ECCV},
  year={2024},
}

@article{chu2025fine,
  title={Fine-Grained Captioning of Long Videos through Scene Graph Consolidation},
  author={Chu, Sanghyeok and Seo, Seonguk and Han, Bohyung},
  journal={arXiv preprint arXiv:2502.16427},
  year={2025}
}

@article{pei2025egothinker,
  title={Egothinker: Unveiling egocentric reasoning with spatio-temporal cot},
  author={Pei, Baoqi and Huang, Yifei and Xu, Jilan and He, Yuping and Chen, Guo and Wu, Fei and Qiao, Yu and Pang, Jiangmiao},
  journal={arXiv preprint arXiv:2510.23569},
  year={2025}
}

@inproceedings{bolya2025perception,
title={Perception Encoder: The best visual embeddings are not at the output of the network},
author={Daniel Bolya and Po-Yao Huang and Peize Sun and Jang Hyun Cho and Andrea Madotto and Chen Wei and Tengyu Ma and Jiale Zhi and Jathushan Rajasegaran and Hanoona Abdul Rasheed and Junke Wang and Marco Monteiro and Hu Xu and Shiyu Dong and Nikhila Ravi and Shang-Wen Li and Piotr Dollar and Christoph Feichtenhofer},
booktitle=NIPS,
year={2025},
}

@article{
oquab2024dinov,
title={{DINO}v2: Learning Robust Visual Features without Supervision},
author={Maxime Oquab and Timoth{\'e}e Darcet and Th{\'e}o Moutakanni and Huy V. Vo and Marc Szafraniec and Vasil Khalidov and Pierre Fernandez and Daniel HAZIZA and Francisco Massa and Alaaeldin El-Nouby and Mido Assran and Nicolas Ballas and Wojciech Galuba and Russell Howes and Po-Yao Huang and Shang-Wen Li and Ishan Misra and Michael Rabbat and Vasu Sharma and Gabriel Synnaeve and Hu Xu and Herve Jegou and Julien Mairal and Patrick Labatut and Armand Joulin and Piotr Bojanowski},
journal={Transactions on Machine Learning Research},
year={2024},
}

@inproceedings{wang2021minilmv2,
  title={Minilmv2: Multi-head self-attention relation distillation for compressing pretrained transformers},
  author={Wang, Wenhui and Bao, Hangbo and Huang, Shaohan and Dong, Li and Wei, Furu},
  booktitle={Findings of the Association for Computational Linguistics: ACL-IJCNLP 2021},
  year={2021}
}

@inproceedings{peirone2025hiero,
  title={HiERO: understanding the hierarchy of human behavior enhances reasoning on egocentric videos},
  author={Peirone, Simone Alberto and Pistilli, Francesca and Averta, Giuseppe},
  booktitle={ICCV},
  year={2025}
}

@inproceedings{
    ravi2025sam,
    title={{SAM} 2: Segment Anything in Images and Videos},
    author={Nikhila Ravi and Valentin Gabeur and Yuan-Ting Hu and Ronghang Hu and Chaitanya Ryali and Tengyu Ma and Haitham Khedr and Roman R{\"a}dle and Chloe Rolland and Laura Gustafson and Eric Mintun and Junting Pan and Kalyan Vasudev Alwala and Nicolas Carion and Chao-Yuan Wu and Ross Girshick and Piotr Dollar and Christoph Feichtenhofer},
    booktitle={ICLR},
    year={2025},
}

@inproceedings{li2023blip,
  title={Blip-2: Bootstrapping language-image pre-training with frozen image encoders and large language models},
  author={Li, Junnan and Li, Dongxu and Savarese, Silvio and Hoi, Steven},
  booktitle={ICML},
  year={2023},
}
}
\appendix
\newpage
\appendix
\section*{Appendix}

\section{The \oursdata dataset} \label{sec:app:sgego}
\begin{figure}[!b]
        \centering
        \includegraphics[width=0.95\textwidth]{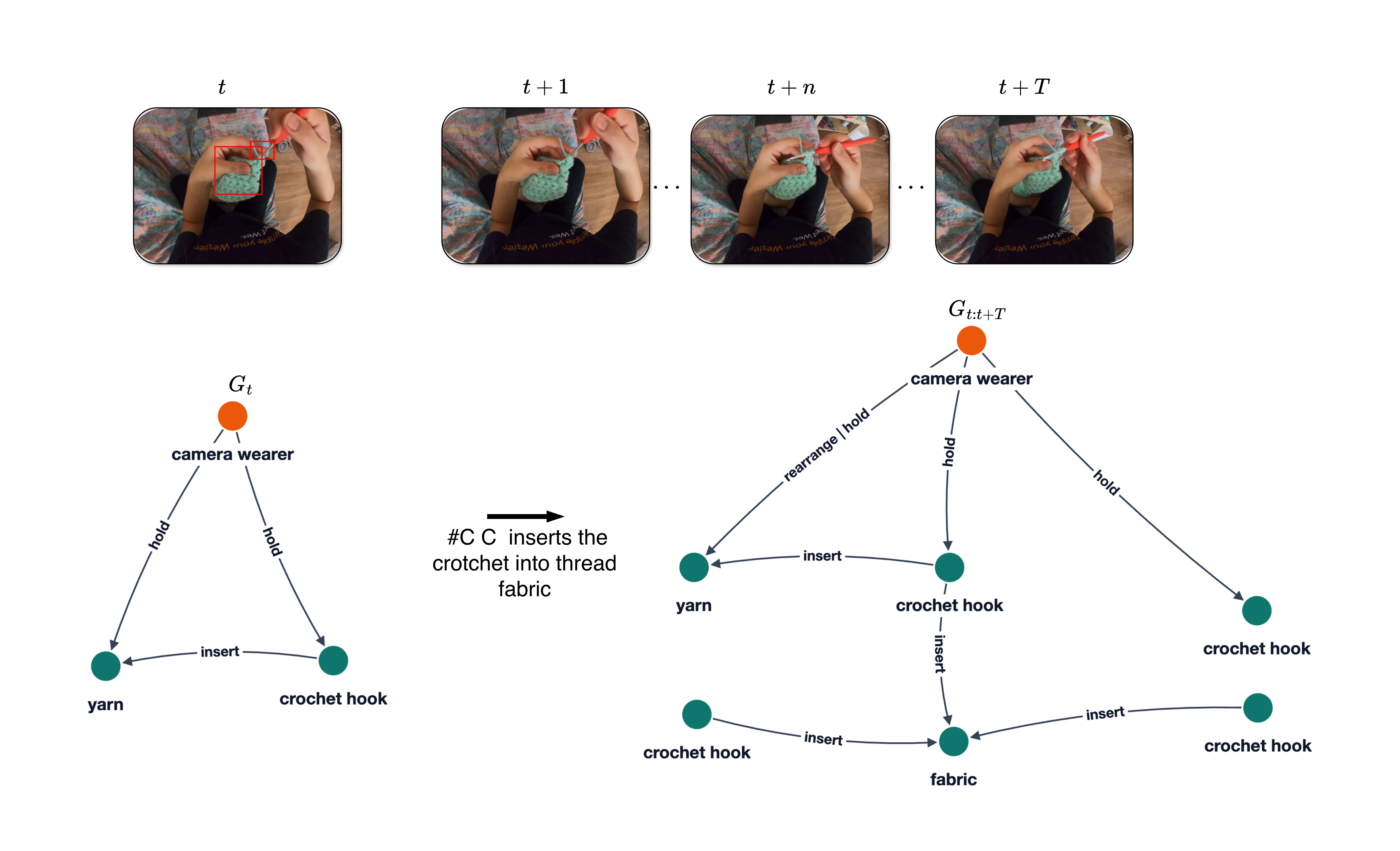}
        \vspace{-1mm}
        \caption{\textbf{Qualitative examples from the \oursdata dataset}. Video UID: \texttt{a1b43a0c-cfaa-44df-adcc-531697ce1f8c}, from second 70.4 to 71.6. Notice that in $G_{t:t+T}$, multiple crochet hook nodes appear because the graph tracks the main actor while knitting, and the crochet hook bounding box changes significantly over time as the action progresses.}

        \label{fig:sub1}
\end{figure}

\begin{figure}[!b]
        \centering
        \includegraphics[width=0.95\textwidth]{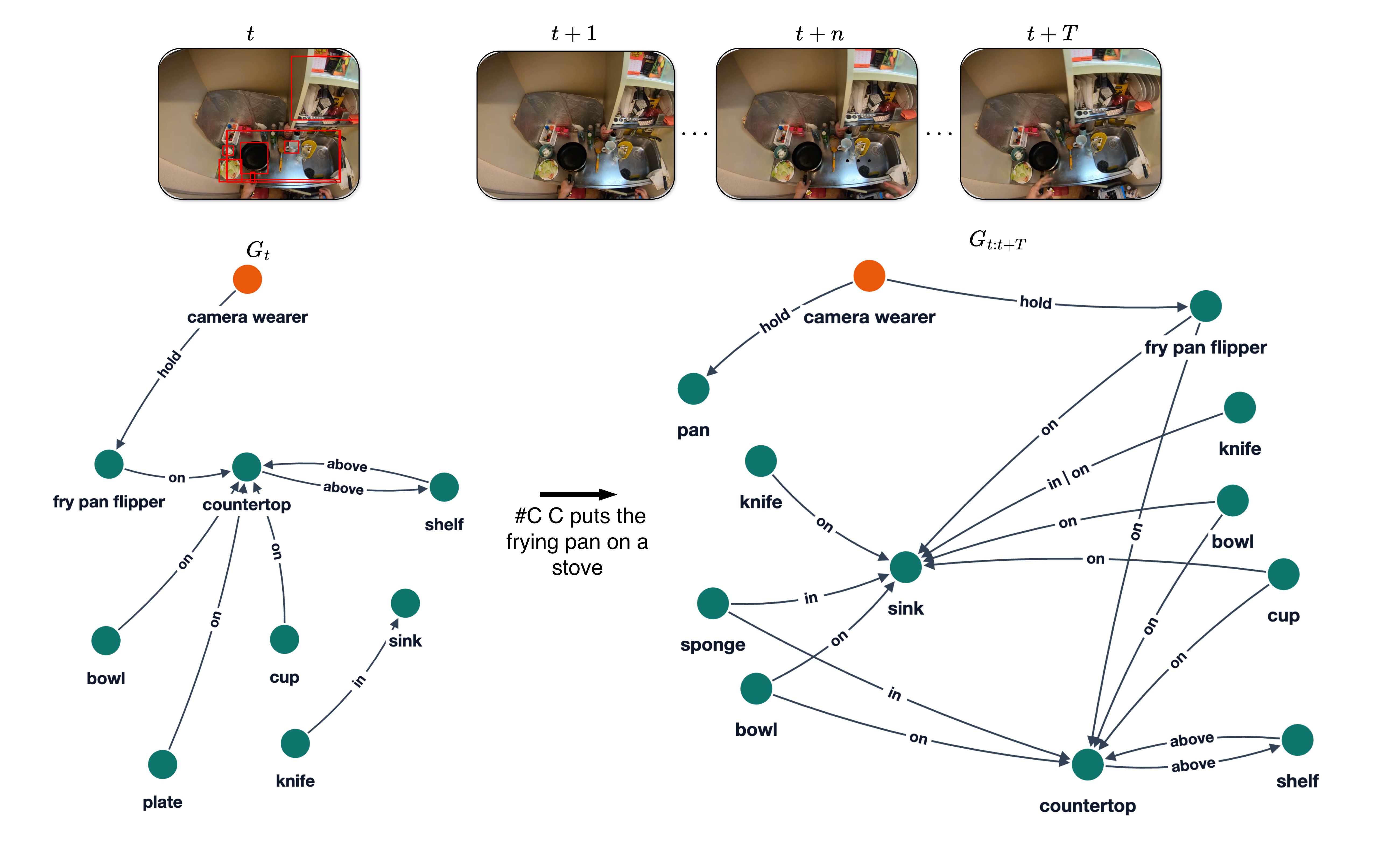}
        \vspace{-0.5mm}
        \caption{\textbf{Qualitative examples from the \oursdata dataset}. Video UID: \texttt{727ffce8-20ec-4111-af26-698eb306e8c7}, from second 74.6 to 75. Interestingly, by aggregating information across the temporal action window, $G_{t:t+T}$ captures a richer representation of the scene than the frame-level graph $G_t$. For example, the two \textit{knife} nodes correspond to distinct knives both present in the environment but not captured in a single frame-level observation.}
        \label{fig:sub2}
\end{figure}

\begin{figure}[!b]
        \centering
        \includegraphics[width=0.95\textwidth]{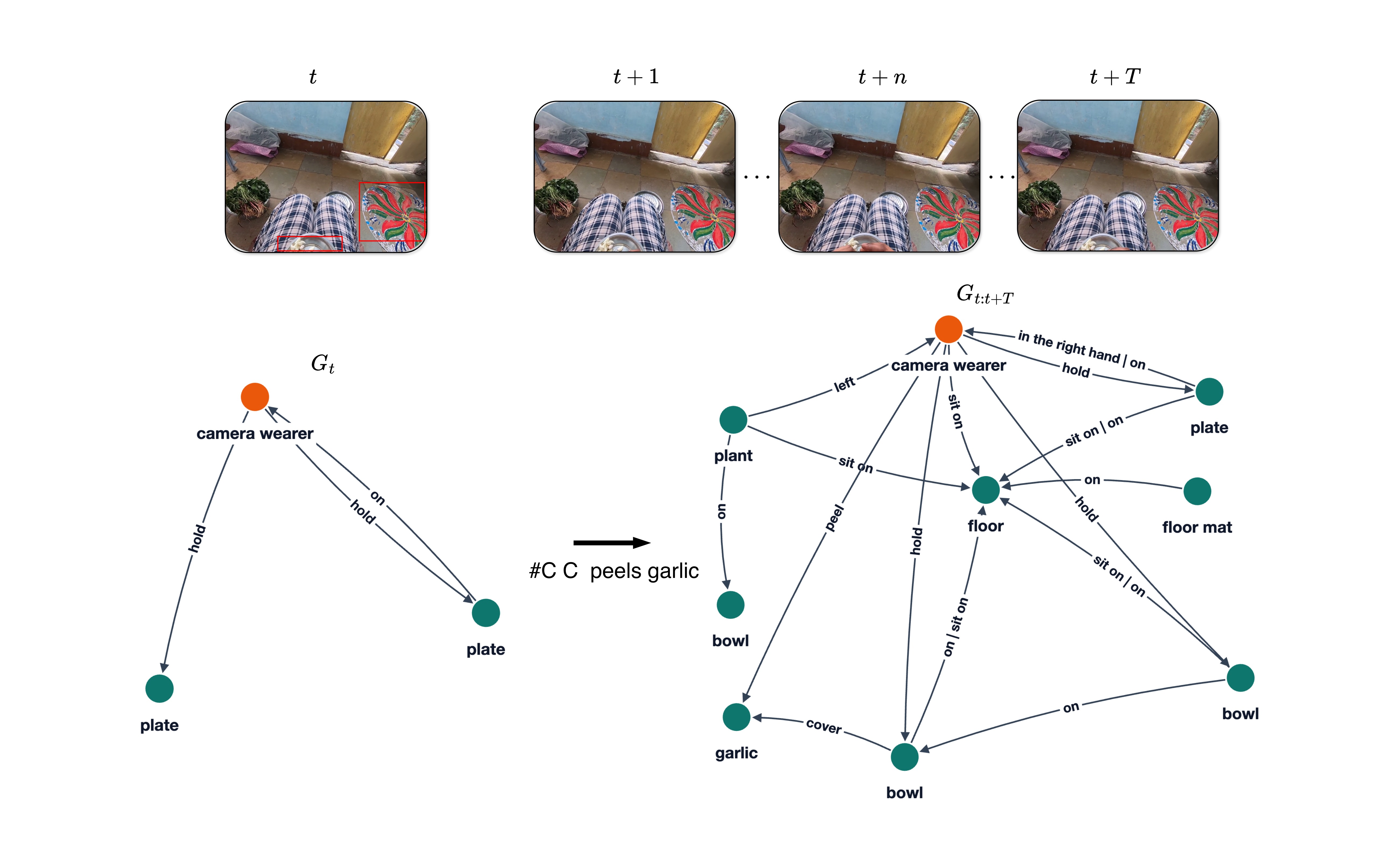}
        \vspace{-1mm}
        \caption{\textbf{Qualitative examples from the \oursdata dataset}. Video UID: \texttt{31d6fe77-da70-42da-8f47-66bb79b9285b}, from second 89.8 to 90.8.
        Notice that the temporal aggregation in $G_{t:t+T}$ enriches the scene representation by incorporating objects and relations observed at different moments of the action sequence. While the initial graph $G_t$ only captures a limited subset of the environment, the consolidated spatio-temporal graph reveals additional objects (\eg, bowls, garlic and plant) together with evolving interactions involving the main actor.}
        \label{fig:sub3}
    \end{figure}

\textbf{\oursdata Examples.} We show some examples from our dataset in Figure~\ref{fig:sub1}, Figure~\ref{fig:sub2} and Figure~\ref{fig:sub3}, reporting the spatial scene graph at the start of the action ($G_t$), the textual narration describing the human action and the final consolidated spatio-temporal scene graph over the entire temporal window~$G_{t:t+T}$.

    \label{fig:qualitatives}

\textbf{\oursdata-Align.} 
The Align dataset consists of approx. 3.8M spatio-temporal graphs from 7297 unique videos. 
Each video is paired with its corresponding textual narration from Ego4D. 
We leverage the EgoClip~\cite{lin2022egocentric} annotations to define a temporal window around each annotated action.
These graphs have $10.97\pm6.58$ nodes and $16.06\pm12.24$ edges on average.

\textbf{\oursdata-Edit.}
The Edit dataset consists of 360k training and 7.2k validation samples. Samples are taken from $7025$ and $182$ unique videos for the train and validation splits, respectively. 
In the \textbf{train split}, the start graphs have $5.44\pm3.42$ nodes and $4.35\pm3.33$ edges on average. The spatio-temporal graphs consolidated over the entire window have $11.68\pm6.50$ nodes and $17.28\pm12.05$ edges on average. 
In the \textbf{validation split}, the start graphs have $5.10\pm3.35$ nodes and $4.05\pm3.24$ edges on average. The spatio-temporal graphs consolidated over the entire window have $12.33\pm7.19$ nodes and $20.36\pm15.49$ edges on average.

\begin{figure}[t]
    \centering
    \begin{subfigure}{0.9\linewidth}
        \centering
        \includegraphics[width=\linewidth]{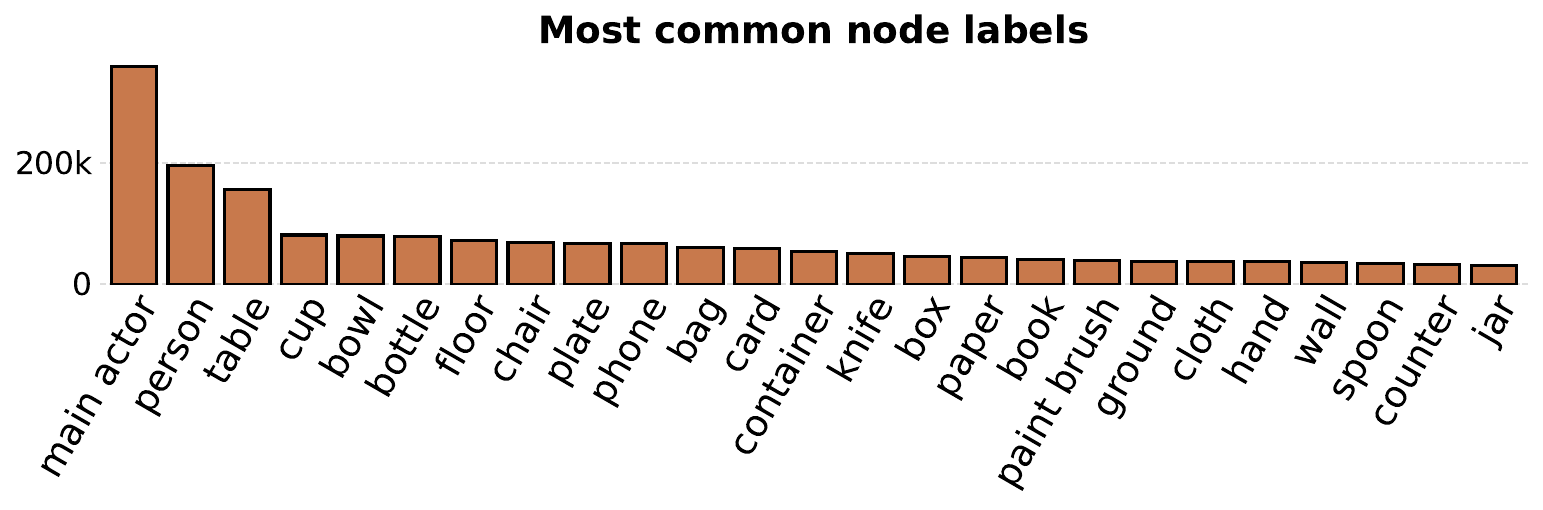}
        \label{fig:sgego_edit_distribution_nodes}
    \end{subfigure}
    \begin{subfigure}{0.9\linewidth}
        \centering
        \includegraphics[width=\linewidth]{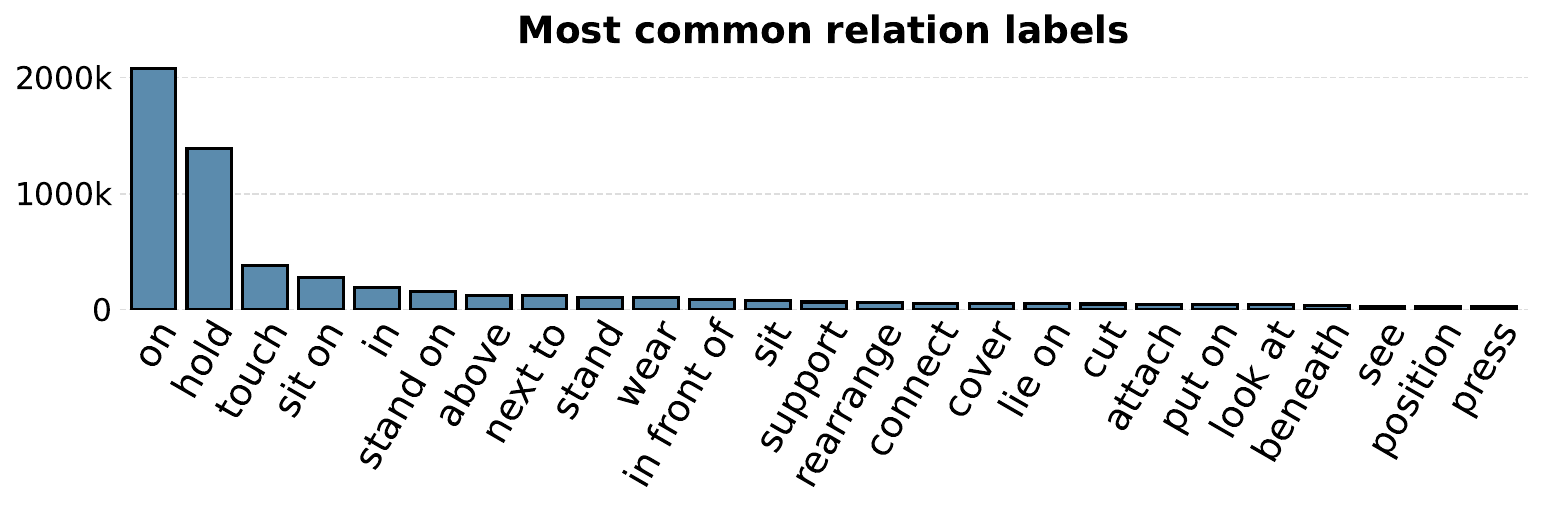}
        \label{fig:sgego_edit_distribution_rels}
        \vspace{-2.5mm}
    \end{subfigure}
    \caption{\textbf{Node and edge label distribution in \oursdata-Edit.} We define a vocabulary of object and relation labels by mining the Ego4D taxonomy and the PVSG relation labels. Following this process, we obtain a set of $N_{obj}=1480$ objects and $N_{rel}=387$ relations. }
    \label{fig:sgego_edit_distribution}
\end{figure}

\textbf{Object and relation vocabularies.} We define a closed set vocabulary of node and edge labels. Specifically, we mine node labels from the Ego4D nouns taxonomy and remove synonyms by clustering the corresponding text embeddings. Similarly, relations are taken from the Ego4D verbs taxonomy and from the relations labels of PVSG~\cite{yang2023panoptic}. Following this process, we obtain a set of $N_{obj}=1480$ objects and $N_{rel}=387$ relations. We report node and edges distributions in Figure~\ref{fig:sgego_edit_distribution}. Note that we label the node relative to the camera wearer, \ie, the human recording the video and performing the main actions, as the ``main actor'', in order to differentiate it with other possible people in the scene, labeled as ``person''.

\subsection{Annotation Pipeline}
For frame-level triplets extraction, we use Qwen3.5-9B~\cite{qwen35}. We prompt the model with one frame at the time, disable thinking and set the generation temperature to $t=0.7$. We report the full prompt used to extract the frame-level triplets in the box "Triplets extraction prompt".
\begin{tcolorbox}[breakable,title=Triplets extraction prompt]\label{prompt}
You are an expert in human-object and object-object interaction analysis. \\
You are given an image corresponding to a frame coming from an egocentric video and your final goal is to generate a good description of the interactions in the image. \\
\\
Output the relations as a list of triplets "subject, predicate, object", one per line. \\
\\
Indications YOU MUST FOLLOW: \\
- Strictly follow the output format required and only include human-object or object-object interactions. For example, do not output "(wall, white)" because it's not an object-object interaction. \\
- Be exhaustive about all the spatial and functional interactions in the scene. \\
- Be specific in the relation types and avoid vague predicates like 'is', 'has' or 'near'. \\
- Refer to the human wearing the egocentric camera as "camera\_wearer", and use "human\_x" for any other people in the scene. \\
- DO NOT repeat the same triplet more than once. \\
- DO NOT include attributes about the objects, such as their color, size, or other visual details. Focus only on the interactions between them. \\
- Include spatial relations between human and objects and between objects (such as "cup\_1, on, table\_1"). \\
- Include functional relations only between the human and the objects in the scene (such as "human\_1, holding, cup\_2"). \\
- DO NOT include functional relations between objects in the scene (such as "cup\_1, used for, drinking"). Use only spatial relations for object-object interactions. \\
- Keep track of multiple instances of the same object category in the scene using numerical indices (e.g., "cup\_1", "cup\_2") and use these indices consistently across all triplets to refer to the same instance. \\
- For objects that are not involved in any interaction, do not describe them. \\
- DO NOT describe what the human is wearing.
\end{tcolorbox}

\section{GLEN Ablations and extended results}\label{sec:app:ablations}

\subsection{Graph-Text Alignment}
\begin{table*}[htbp]
    \centering
    \caption{\textbf{Ablations on EgoMCQ.} We compare different configurations of \ours using the Graph-Text Contrastive Alignment (\textbf{GTCA}) and Graph-Text Matching (\textbf{GTM}) objectives. For prediction, we can use either the GTCA head (trained with the contrastive objective), the GTM head (trained for graph-text matching) or an ensemble of the two heads. We indicate with $n$ the number of negative samples for each anchor in the batch.}
    \label{tab:egomcq_ablation}
    \vspace{2mm}
    \begin{tabular}{@{}p{0.49\textwidth}@{\hspace{0.02\textwidth}}p{0.49\textwidth}@{}}
        \begin{minipage}[t]{0.5\textwidth}
            \centering
            \scriptsize
            \setlength{\tabcolsep}{2.5pt}
            \begin{tabularx}{0.9\linewidth}{Xcc}
                \toprule
                \textbf{Configuration}  & \textbf{EgoMCQ Inter} & \textbf{EgoMCQ Intra} \\
                \midrule
                GTCA ($n = 1$) & 89.3 & 53.2 \\
                GTCA ($n = 3$) & 90.6 & 53.9\\
                \midrule
                GTCA ($n = 1$) + GTM & 89.6 & 53.3\\
                GTCA ($n = 3$) + GTM & \textbf{91.0} & 54.4\\
                \midrule
                GTCA ($n = 1$) + GTM$^\dagger$ & 89.4 & 53.7\\
                GTCA ($n = 3$) + GTM$^\dagger$ & \textbf{91.0} & \textbf{54.9}\\
                \bottomrule
            \end{tabularx}
        \end{minipage}
         &
        \begin{minipage}[t]{0.5\textwidth}
            \centering
            \scriptsize
            \setlength{\tabcolsep}{3.5pt}
            \begin{tabularx}{0.9\linewidth}{Xcc}
                \toprule
                \textbf{Mode}  & \textbf{EgoMCQ Inter} & \textbf{EgoMCQ Intra} \\
                \midrule
                GTCA head & 91.0 & 54.9 \\
                GTM$^\dagger$ head & 80.8 & 54.0 \\
                GTCA + GTM$^\dagger$ heads  & \textbf{91.2} & \textbf{56.2} \\
                \bottomrule
            \end{tabularx}
        \end{minipage} \\
    \end{tabular}
\end{table*}
We report in Table~\ref{tab:egomcq_ablation} (left) the impact of different training objectives for the Graph-Text Alignment task on EgoMCQ. Here, $n$ denotes the number of negatives sampled for each anchor within the batch. 
We observe that hard negative mining significantly improves graph-text alignment performance. Since scene graphs provide a high-level abstraction of the evolving scene, this strategy encourages the model to distinguish between actions occurring in visually similar scenes while capturing different underlying actions.
The GTM objective further helps the model better distinguish between similar scene graphs. The GTM$^\dagger$ variant further restricts sampling to the same video.

For the evaluation in Table~\ref{tab:egomcq_ablation} (right), we use either the cross-modal similarity between the graph and text embeddings (GTCA head), the output of the GTM head or a combination of the two.

\subsection{Activity-conditioned graph edit forecasting}
We show in Table~\ref{tab:agef_abl} the impact of the number of node queries on the A-GEF task. Increasing the number of queries allows the model to specialize each query on different objects which is particularly helpful on the large consolidated graphs from \oursdata-Edit.

\subsection{Long-Horizon Reasoning on EXPLORE-Bench}\label{sec:app:explore}
We present the full results for Explore-Bench in Table~\ref{tab:explore_app}.
\begin{table}[htbp]
    \caption{\textbf{Activity-conditioned graph edit forecasting (A-GEF) on \oursdata}.}
    \label{tab:agef_abl}
    \centering
    \scriptsize
    \begin{tabular}{llllllll}
    \toprule

        \multirow{2}{*}{\bf Num queries}  & \multicolumn{3}{c}{\bf Triplet Recall} \\ 

           & R@20 & R@50 & R@100   \\
        \midrule
        $G_t$ (\emph{static})  & 25.94 & 25.94 & 25.94  \\
64  & 36.84 & 44.06 & 46.61\\
72 & 36.93 & 44.63 & 47.55\\
100 & 36.73 & 45.30 & 49.38\\
128 & 36.69	& 45.39	& 49.91\\
        \bottomrule
    \end{tabular}
\end{table}
\begin{table}[H] 
\caption{\textbf{Evaluation results on EXPLORE-Bench.} \texttt{Short}, \texttt{Medium} and \texttt{Long} mean the subsets with short (11-99), medium (100-199) and long (200-694) atomic-action sequences, respectively. $^\dagger$ Reproduced results using the official protocol.
${\#}$ and $*$ represent the non-thinking and thinking mode, respectively.
}
\label{tab:explore_app}
\centering
\renewcommand{\arraystretch}{1.1} 
\setlength{\tabcolsep}{2.5pt}
\resizebox{\textwidth}{!}{
\begin{tabular}{l|cccc|cccc|cccc|cccc}
\toprule
\multirow{2}{*}{\textbf{Methods}} & \multicolumn{4}{c|}{\texttt{Short}} & \multicolumn{4}{c|}{\texttt{Medium}} & \multicolumn{4}{c|}{\texttt{Long}} & \multicolumn{4}{c}{\texttt{Full}} \\
 & {$\bm{S_{obj}}$} & {$\bm{S_{att}}$} & {$\bm{S_{rel}}$} & {$\bm{S_{uni}}$} & {$\bm{S_{obj}}$} & {$\bm{S_{att}}$} & {$\bm{S_{rel}}$} & {$\bm{S_{uni}}$} & {$\bm{S_{obj}}$} & {$\bm{S_{att}}$} & {$\bm{S_{rel}}$} & {$\bm{S_{uni}}$} & {$\bm{S_{obj}}$} & {$\bm{S_{att}}$} & {$\bm{S_{rel}}$} & {$\bm{S_{uni}}$} \\ 
\midrule
\multicolumn{17}{c}{\textit{Proprietary Multimodal Foundation Models (API)}} \\ \midrule
GPT-5.2-Chat & 59.91 & 1.74 & 2.70 & 48.71 & 59.88 & 1.67 & 2.65 & 47.85 & 58.06 & 1.65 & 2.61 & 46.91 & 59.69 & 1.70 & 2.67 & 48.19 \\
Gemini-3-Flash & 60.31 & 1.84 & 2.78 & 50.27 & 59.33 & 1.76 & 2.71 & 48.84 & 58.27 & 1.72 & 2.69 & 48.18 & 59.72 & 1.80 & 2.75 & 49.51 \\
Gemini-3-Pro & 61.29 & 1.81 & 2.77 & 50.11 & 60.99 & 1.75 & 2.74 & 49.44 & 59.17 & 1.70 & 2.70 & 48.31 & 60.94 & 1.77 & 2.75 & 49.66 \\ \midrule
\multicolumn{17}{c}{\textit{Open-source Non-thinking Multimodal Foundation Models}} \\ \midrule
Qwen2.5-VL-3B-Instruct & 48.17 & 1.36 & 2.27 & 39.70 & 44.19 & 1.22 & 2.17 & 36.92 & 39.92 & 1.09 & 2.08 & 34.28 & 45.78 & 1.28 & 2.21 & 38.07 \\ Qwen2-VL-7B-Instruct & 47.64 & 1.34 & 2.23 & 39.14 & 46.04 & 1.27 & 2.16 & 37.73 & 43.79 & 1.22 & 2.15 & 36.73 & 46.62 & 1.30 & 2.20 & 38.35 \\
Keye-VL-1.5-8B$^{\#}$ & 49.70 & 1.34 & 2.24 & 39.78 & 49.49 & 1.34 & 2.24 & 39.69 & 46.15 & 1.25 & 2.17 & 37.68 & 49.23 & 1.33 & 2.23 & 39.51 \\
LLaVA-OneVision-1.5-4B & 51.15 & 1.42 & 2.33 & 41.37 & 49.17 & 1.33 & 2.28 & 39.85 & 46.35 & 1.28 & 2.26 & 38.64 & 49.88 & 1.37 & 2.31 & 40.50 \\
Qwen3-VL-2B-Instruct & 52.18 & 1.48 & 2.55 & 43.77 & 49.03 & 1.37 & 2.45 & 41.38 & 43.27 & 1.18 & 2.28 & 37.34 & 50.02 & 1.40 & 2.48 & 42.17 \\
Qwen2.5-VL-7B-Instruct & 52.80 & 1.52 & 2.46 & 43.58 & 51.25 & 1.46 & 2.41 & 42.32 & 47.78 & 1.34 & 2.36 & 40.22 & 51.67 & 1.48 & 2.43 & 42.74 \\
LLaVA-OneVision-1.5-8B & 53.25 & 1.54 & 2.51 & 44.13 & 51.21 & 1.47 & 2.44 & 42.63 & 47.62 & 1.38 & 2.41 & 40.83 & 51.87 & 1.49 & 2.47 & 43.21 \\
Ovis2.5-2B$^{\#}$ & 55.65 & 1.60 & 2.56 & 45.64 & 51.70 & 1.44 & 2.43 & 42.49 & 47.96 & 1.33 & 2.35 & 40.13 & 53.33 & 1.51 & 2.49 & 43.86 \\
InternVL3.5-2B & 57.51 & 1.64 & 2.62 & 46.80 & 55.35 & 1.53 & 2.50 & 44.60 & 52.40 & 1.42 & 2.41 & 42.32 & 56.14 & 1.58 & 2.55 & 45.48 \\
InternVL3.5-8B & 57.85 & 1.65 & 2.62 & 47.00 & 55.77 & 1.56 & 2.55 & 45.30 & 53.30 & 1.51 & 2.51 & 43.99 & 56.57 & 1.60 & 2.58 & 46.04 \\
Qwen3.5-9B-Instruct$^\dagger$ & 58.73 & - & 2.74 & - & 56.20 & - & 2.63 & - & 51.21 & - & 2.55 & - & 56.95 & - & 2.68 & - \\
MiMo-VL-7B-RL-2508$^{\#}$ & 56.13 & 1.67 & 2.60 & 46.53 & 55.82 & 1.64 & 2.59 & 46.14 & 53.42 & 1.58 & 2.54 & 44.68 & 55.71 & 1.65 & 2.59 & 46.18 \\
MiniCPM-V-4.5 (8B)$^{\#}$ & 58.86 & 1.70 & 2.63 & 47.67 & 57.45 & 1.64 & 2.58 & 46.49 & 54.69 & 1.56 & 2.55 & 44.95 & 57.87 & 1.66 & 2.60 & 46.93 \\
Ovis2.5-9B$^{\#}$ & 59.51 & 1.74 & 2.72 & 48.85 & 57.84 & 1.65 & 2.63 & 47.01 & 53.83 & 1.55 & 2.50 & 44.32 & 58.26 & 1.68 & 2.66 & 47.66 \\
Qwen3-VL-8B-Instruct & 61.34 & 1.88 & 2.84 & 51.23 & 60.78 & 1.85 & 2.81 & 50.64 & 56.83 & 1.73 & 2.71 & 48.00 & 60.63 &1.85 & 2.82 & 50.65 \\ \midrule
\multicolumn{17}{c}{\textit{Open-source Thinking Multimodal Foundation Models}} \\ \midrule
Ovis2.5-2B$^*$ & 51.10 & 1.40 & 2.43 & 42.02 & 44.52 & 1.17 & 2.20 & 36.89 & 40.70 & 1.04 & 2.10 & 34.25 & 47.50 & 1.28 & 2.31 & 39.25 \\
Qwen3-VL-2B-Thinking & 59.23 & 1.47 & 2.63 & 46.13 & 57.02 & 1.19 & 2.53 & 42.79 & 48.82 & 1.02 & 2.30 & 37.81 & 57.26 & 1.32 & 2.56 & 43.97 \\
Ovis2.5-9B$^*$ & 55.92 & 1.61 & 2.59 & 45.95 & 53.79 & 1.51 & 2.50 & 44.03 & 49.71 & 1.39 & 2.41 & 41.48 & 54.44 & 1.55 & 2.54 & 44.74 \\
Keye-VL-1.5-8B$^*$ & 56.58 & 1.58 & 2.55 & 45.64 & 56.54 & 1.54 & 2.53 & 45.14 & 53.13 & 1.62 & 2.43 & 44.09 & 56.18 & 1.57 & 2.53 & 45.28 \\
MiMo-VL-7B-RL-2508$^*$ & 57.32 & 1.64 & 2.56 & 46.26 & 56.85 & 1.59 & 2.53 & 45.63 & 56.58 & 1.57 & 2.50 & 45.17 & 57.06 & 1.61 & 2.54 & 45.90 \\
MiniCPM-V-4.5 (8B)$^*$ & 58.87 & 1.74 & 2.67 & 48.25 & 57.79 & 1.69 & 2.63 & 47.33 & 54.16 & 1.54 & 2.52 & 44.46 & 57.95 & 1.70 & 2.64 & 47.49 \\
GLM-4.6V-Flash (9B) & 60.70 & 1.78 & 2.71 & 49.32 & 58.44 & 1.70 & 2.65 & 47.67 & 54.95 & 1.60 & 2.60 & 45.75 & 59.22 & 1.73 & 2.67 & 48.31 \\
Step3-VL-10B & 61.32 & 1.76 & 2.78 & 49.87 & 60.06 & 1.62 & 2.71 & 48.01 & 59.11 & 1.60 & 2.64 & 47.09 & 60.61 & 1.69 & 2.74 & 48.87 \\
Qwen3-VL-8B-Thinking & 63.77 & 1.91 & 2.85 & 52.08 & 62.61 & 1.81 & 2.78 & 50.56 & 58.02 & 1.64 & 2.63 & 47.07 & 62.70 & 1.84 & 2.80 & 50.96 \\ \midrule
\multicolumn{17}{c}{\textit{Embodied/Egocentric Multimodal Models}} \\ \midrule
Embodied-Reasoner (7B) & 39.74 & 1.00 & 1.96 & 32.65 & 39.69 & 0.99 & 1.96 & 32.56 & 37.22 & 0.97 & 2.02 & 32.23 & 39.44 & 1.00 & 1.97 & 32.57 \\
EgoThinker (7B) & 48.22 & 1.36 & 2.23 & 39.39 & 45.88 & 1.28 & 2.19 & 37.89 & 42.42 & 1.16 & 2.08 & 35.36 & 46.71 & 1.31 & 2.20 & 38.38 \\
\midrule
\multicolumn{17}{c}{\textit{Graph Edits}} \\ \midrule
\rowcolor{lightgreenbg}
\ours & \textbf{66.12} & - & 2.67 & - & \textbf{66.71} & - & 2.73 & - & 59.37 & - & 2.67 & - &\textbf{65.59} & - & 2.69 & -  \\
\bottomrule
\end{tabular}%
}
\end{table}


\end{document}